\documentclass{artificial-life}

\usepackage{amsmath}
\usepackage{amssymb}

\usepackage[style=apa,natbib=true]{biblatex}
\title{Evolutionary Brain-Body Co-Optimization Consistently Fails to Select for Morphological Potential}

\auth{
    Alican Mertan \affil{1}, 
    Nick Cheney \affil{1}
}

\corresponding{Alican Mertan}{alican.mertan@pm.me}

\affiliations{
    \item Neurobotics Lab, University of Vermont, USA
}

\abs{
Brain-body co-optimization remains a challenging problem. To understand and overcome its challenges, we exhaustively map a morphology-fitness landscape: we train controllers for each morphology in a design space of 1,305,840 voxel-based soft robots. We show that this design space constitutes a good model for studying brain-body co-optimization and that our mapping roughly captures its landscape. Complete knowledge of the landscape lets us analyze how evolutionary co-optimization algorithms unfold. We find that the tested algorithms cannot consistently find near-optimal solutions: the search, at times, gets stuck on morphologies one mutation away from better ones, because it regularly undervalues individuals with newly mutated bodies and eliminates promising morphologies. On the other hand, co-optimizing morphology and control creates useful goal-switching, yielding morphology-controller pairs whose performance cannot be reached by optimizing the controller alone for a fixed morphology. Together, these results ground trends in the literature and offer insights for future work.
}


\begin{document}

\coverpage           %
\doublespacing       %


\section{Introduction}

Brain-body co-optimization refers to the process of optimizing both the control and design of a robot for a given task. While the origins of brain-body co-optimization go back more than 30 years~\citep{sims_evolving_1994}, there has been a recent increase in research interest~\citep{liu_embodied_2025,wang2025brainbody}. Especially with the advent of soft robots consisting of materials capable of highly complex and unintuitive behavior~\citep{hiller_automatic_2012}, the need for automated design and control has increased greatly. Moreover, studies of embodied intelligence have shown how specialized designs for a given task and environment can improve the efficacy and efficiency of robot designs, find more robust solutions, or find ingenious designs that would be difficult to design by hand~\citep{pfeifer_how_2007,bhatia_evolution_2021}.

Brain-body co-optimization, however, still remains a challenging problem despite great interest from the research community. 
Previous studies have highlighted a key issue, namely the premature convergence of morphology~\citep{joachimczak_artificial_2016,cheney_difficulty_2016,mertan_investigating_2024} -- the evolutionary co-optimization of brain and body fails to fully optimize the morphology.
In this work, our objective is to further improve our understanding of the problem and facilitate research in this domain. Previous literature offers a general approach to such a problem, where low-level features of fitness landscapes are estimated to characterize them~\citep{mersmann2010benchmarking,mersmann2011exploratory}. Here, to gain a deeper understanding of a specific optimization problem, we propose a different, more computationally demanding approach. Inspired by the success of the NAS-Bench-101 public architecture dataset and its derivatives in fostering more research and benefiting the neural architecture search community~\citep{ying2019bench,dong_nas-bench-201_2020}, we set out to create a similar dataset where we measure the true fitness of each morphology in a given design space, building an exhaustive morphology-fitness landscape.

To map the morphology-fitness landscape exhaustively, we train controllers for each feasible morphology in a design space of 1,305,840 distinct voxel-based soft robot morphologies in simulation engine Evogym~\citep{bhatia_evolution_2021}, constrained by a computational budget. We use the best performance achieved during the training as an estimate of their best possible performance. We investigate the validity and significance of the data and conclude that it provides a good test bed to study the brain-body co-optimization problem. We believe that this data has the potential to improve our understanding by providing complete access to the search space. 

To make an initial attempt to use this data to gain insights into the brain-body co-optimization problem, we investigate how three commonly used evolutionary optimization algorithms, namely age-fitness Pareto optimization (AFPO)~\citep{schmidt_age-fitness_2010}, MAP-Elites~\citep{mouret_illuminating_2015}, and morphological innovation protection (MIP)~\citep{cheney_scalable_2018}, work on this morphology-fitness landscape. Having access to the complete morphology-fitness landscape allows us to demonstrate that the evolutionary optimization algorithms tested often fail to find near-optimal solutions. Most strikingly, we show that all of the tested algorithms, at times, get stuck in morphologies that are one mutation away from a morphology with better performance. We show that they frequently eliminate morphologies with better performance due to their inability to estimate true potential. Although previous research has provided evidence suggesting this phenomenon -- fragile co-adaptation~\citep{joachimczak_artificial_2016,cheney_difficulty_2016,mertan_investigating_2024}, having access to the full morphology-fitness landscape enables us to clearly illustrate fragile co-adaptation and its implications. 

Our investigation also documents cases where the performance of a morphology-controller pair found through evolutionary brain-body co-optimization cannot be reached by optimizing only the controller while keeping the morphology fixed. This suggests that a continuous control problem can have a deceptive optimization landscape~\citep{lehman2011novelty,lehman_abandoning_2011} where optimizing a controller directly for the target morphology gets stuck in a local optimum. On the other hand, evolutionary brain-body co-optimization might create useful goal-switching~\citep{nguyen2015innovation} -- in the lineage of a morphology-controller pair, a controller is selected for its performance on different morphologies at different times in evolution, which might be helping the optimization process to avoid deceptive local optima. 

Lastly, we review the existing literature in light of our findings, suggesting potential future work directions, both to make use of the provided data and to overcome challenges of brain-body co-optimization.

In summary, our contributions are as follows.
\begin{itemize}
    \item We provide a dataset of a complete morphology-fitness landscape -- estimated true fitness values of all feasible voxel-based soft robots in a design space. We provide evidence that the mapped morphology-fitness landscape constitutes a good model to study the co-optimization problem. We hope that this data facilitates further research in this domain\footnote{Data and code are available at \href{https://github.com/mertan-a/morphology-fitness-landscape}{our repository.}} (Fig.~\ref{fig:ranking_correlation},~\ref{fig:full-fitness-dist},~\ref{fig:body_analysis}).
    \item We investigate how three different evolutionary brain-body co-optimization algorithms, namely age-fitness Pareto optimization~\citep{schmidt_age-fitness_2010}, MAP-Elites~\citep{mouret_illuminating_2015}, and morphological innovation protection~\citep{cheney_scalable_2018}, work by using complete knowledge of the morphology-fitness landscape, providing the most concrete evidence to date (Section~\hyperlink{sect:bb-coop}{Brain-body co-optimization}).
    \item We show that ranking high-performing individuals correctly requires extensive controller training and hypothesize that this need, combined with fragile co-adaptation, hinders the optimization of morphologies in evolutionary brain-body co-optimization by creating dynamics leading to first-mover advantages (Fig.~\ref{fig:ranking_correlation},~\ref{fig:coop-fitness-change},~\ref{fig:coop-elimination}).
    \item We show that search, at times, stagnates at points in the morphology-fitness landscape that are not even a local maximum. We provide evidence to show that evolutionary brain-body co-optimization algorithms consistently undervalue the fitness of offspring with newly mutated bodies, eliminating promising morphologies (Fig.~\ref{fig:rc-coop-vs-updated},~\ref{fig:rc-coop-vs-updated-lm-analysis},~\ref{fig:coop-rc-vs-ms-rc},~\ref{fig:coop-fitness-change},~\ref{fig:coop-elimination}).
    \item We document cases where optimizing a controller from scratch for a target morphology cannot reach the performance of a controller found through evolutionary brain-body co-optimization for the same morphology, demonstrating the benefits of co-optimizing morphology and control together (Fig.~\ref{fig:coop-vs-control},~\ref{fig:coop-vs-control-detailed}).  
    \item We use our findings to ground trends in the literature and identify promising research directions (Section~\hyperlink{sect:discussion}{Discussion}).
    
\end{itemize}

An earlier version of this work was presented at the 2025 ALIFE conference~\citep{mertan2025evolutionary}. This investigation extends that foundation in two important directions. First, it broadens the methodological scope by incorporating morphological innovation protection (MIP)~\citep{cheney_scalable_2018} -- an evolutionary brain-body co-optimization algorithm specifically designed to combat fragile co-adaptation. Second, it introduces a novel analysis of co-optimized solutions, revealing unique goal-switching advantages of brain-body co-optimization that have not been previously documented.

\section{Methodology}

In this section, we outline the general experimental setting that serves as the foundation for our subsequent morphology-fitness landscape construction and evolutionary brain-body co-optimization analysis.

\paragraph{Simulation Engine} We use Evogym~\citep{bhatia_evolution_2021} as a simulation engine, which is a 2D voxel-based soft body simulator. Each voxel is modeled by 4 point masses and 6 springs connecting them.

\paragraph{Brain-body co-optimization} The problem we are interested in analyzing is the optimization of control and morphology, that is, finding the best-performing morphology-controller pair for a given environment and task.

\paragraph{Task and fitness} The task we consider in this work is a locomotion task. The environment consists of a flat surface and a target position that robots are tasked to reach. The environment runs until the robot reaches the end of the terrain or for a maximum of 500 time steps.   
We use a modified version of the fitness described in~\citep{bhatia_evolution_2021}, which rewards the robot's progress in the x-direction. We apply an additional small penalty at each time step to facilitate the selection of robots that move as fast as possible. This establishes a fitness scale where completely failing to move yields a fitness of -5, reaching the target at the final time step yields roughly +1, and scores above +1 reflect robots that reach the target before the maximum time step is reached. 

\paragraph{Morphology and its representation} Each robot consists of voxels that are placed on a 2D grid. Designing the morphology of a robot is simply determining the existence/absence of a voxel in a grid location and determining its material properties. The simulation defines 4 kinds of voxels. Two types of voxels are referred to as passive voxels, as they are not under direct control of the controller and can be soft (gray) or rigid (black). The other two types of voxels are referred to as active voxels, and they can be actuated horizontally (orange) or vertically (teal) by the controller. For a morphology to be considered viable, it needs to have at least three active voxels. While the simulation engine requires a baseline of two, we set the threshold to three to encourage a higher baseline of actuation power for locomotion, and to help constrain the total number of morphologies in our design space.

\paragraph{Control and controller representation} 
The controller is modeled by a neural network (a single-layer network with $32$ hidden neurons) that observes proprioceptive data provided by the simulation engine~\citep{bhatia_evolution_2021}. Specifically, the observation vector consists of the position of each point mass relative to the robot's center of mass ($\in \mathbb{R}^{2n}$ where $n$ is the number of point masses and $n = 16$ for the largest robot) and the velocity of the robot's center of mass ($\in \mathbb{R}^2$). To make the controller network compatible with different morphologies, the observation vectors of smaller robots are zero-padded to match the observation vectors of the largest robot. The controller controls the robot by outputting a multiplier for the resting length of the springs ($\in [0.6, 1.6]$), which is resolved by the simulation engine and results in a change in the area for the voxels. The controller is queried every 5\textsuperscript{th} timestep and the last action is repeated for the remaining timesteps to prevent high-frequency dynamics. In practice, the controller outputs a vector of length $9$ (one action per voxel for the biggest robot), and the actions corresponding to non-existing voxels in smaller robots or passive voxels are discarded. In this configuration, the controller network contains a total of 1417 parameters.

\paragraph{Evolution} In addressing the issue of simultaneously optimizing the brain and body, each individual within a population comprises a morphology and a controller. Throughout evolution, offspring are generated by introducing mutations in either the controller or the morphology. When applied to morphology, the mutation operator alters one voxel. For controllers, the mutation operator involves adding a noise vector drawn from a normal distribution to the neural network's weights. Crossover is not considered in this work.

\section{Initial Morphology-Fitness Landscape}\label{sec:imfl}

To understand how current optimization algorithms work on the brain-body co-optimization algorithm and to improve them, we aim to analyze the underlying morphology-fitness landscape and how the algorithms perform optimization on it. To this end, we run an optimization algorithm exhaustively on a 3-by-3 morphology space to determine their true fitness values -- the best possible performance of a robot given the optimal control for the locomotion task on a flat surface. 

The morphology space is small -- we only consider robots that fit into a 3-by-3 grid, having 9 voxels at most. However, even this small space contains 1,305,840 distinct viable morphologies. In practice, to keep computational cost tractable, we run 300 generations of age-fitness Pareto optimization~\citep{schmidt_age-fitness_2010} with a population size of 20, to optimize the weights of a single hidden layer neural network for each viable morphology that exists in the 3-by-3 morphology space. We treat the best controller found at the end of 300 generations as the estimated optimal controller, and the fitness of the morphology with the estimated optimal controller is treated as the estimated true fitness of that morphology. 

\paragraph{Does this data capture the structure of the morphology-fitness landscape well?}
Since evolutionary brain-body co-optimization algorithms that we test in this work (see Section~\hyperlink{sect:bb-coop}{Brain-body Co-optimization}) only rely on the ordinal relations between fitness values of individuals, as opposed to numerical values, a good match between estimated true fitness values and true fitness values is not essential. Instead, capturing the ranking of morphologies correctly is sufficient. To see if 300 generations of controller optimization were enough to capture the ranking of morphologies correctly, we use the convergence to a particular ranking of morphologies as a proxy. Intuitively, if the ranking of morphologies changes frequently after each controller optimization step, it suggests that more controller optimization is needed. However, if rankings are not changing as controllers are optimized, it indicates that the structure of the morphology-fitness landscape is captured. To this end, we examine how the ranking of morphologies changes over 300 generations of controller optimization. We consider the ranking of all morphologies in the search space (top 100\%), as well as the top 50, 5, and 1\% of morphologies in terms of their estimated true fitness.  

\begin{figure}
    \centering
    \includegraphics[width=0.9\linewidth]{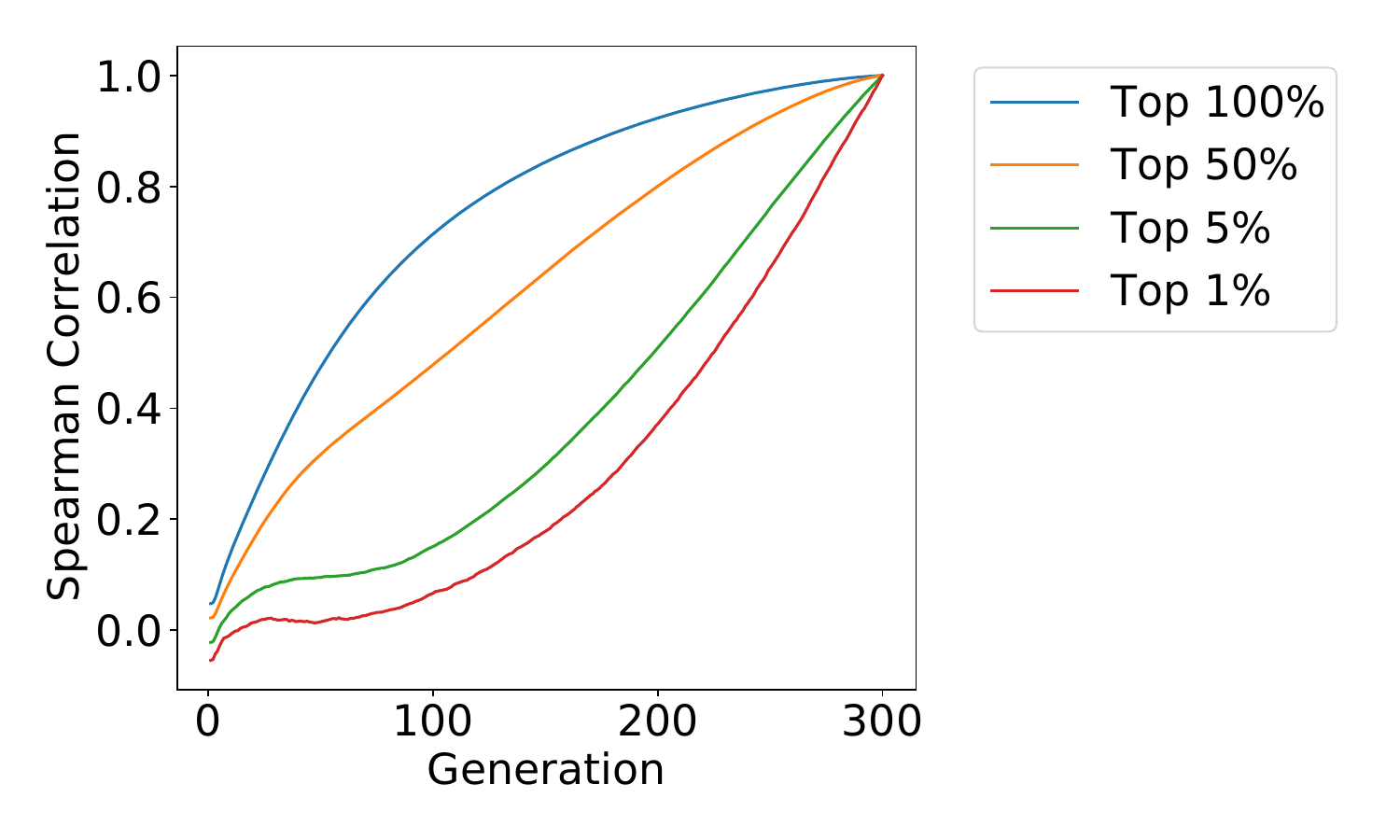}
    \caption{ Plots of the correlation of fitness values of each generation, compared to the final generation's fitness values during controller optimization.
    We measure the correlation in the whole morphology space (Top 100\%) as well as the top 50, 5, and 1\% of morphologies, focusing more and more on the rankings of high-fitness individuals. 
    The correlations are generally increasing as the optimization takes place, demonstrating that rankings are getting consistent over time with the final ranking of the morphologies.
    As we exclude low-fitness individuals from this analysis and focus more on the top-ranked high-fitness individuals, we see that the flat region at the end with the Spearman score of 1 disappears. This suggests that while the dataset captures the landscape roughly, it is noisy when it comes to ranking high-fitness morphologies among themselves. 
    Similarly, as we exclude low-fitness individuals and focus on top-ranked individuals, we see the appearance of a flat region around the Spearman score of 0. This suggests that the selection between top-ranked individuals would be random early on in an evolutionary process. This result has important implications, as it means that evolutionary algorithms can not make informed selections among good morphologies early in evolution.
    }
    \label{fig:ranking_correlation}
\end{figure}

Fig.~\ref{fig:ranking_correlation} illustrates the correlation between the morphology rankings in each generation compared to the final generation's rankings. The flat region towards the end of evolutionary optimization, with a Spearman score of 1, shows that the ranking of morphologies is not changing and is consistent with the final ranking, indicating that the controller optimization has converged. In contrast, the lack of flat regions indicates that more controller optimization can change the ranking of morphologies. The Top 100\% results illustrate that the overall rankings are converging, indicating that the data roughly capture the morphology-fitness landscape. However, as we exclude low-performing morphologies from this analysis and focus more on high-performing morphologies, the flat region disappears, indicating that more controller optimization is required to accurately rank high-performing morphologies between themselves.

Similarly, the flat region around the Spearman score of 0 in early generations indicates that the ranking at that generation does not correlate with the final ranking, meaning that the selection based on those fitness values would be uninformed. Interestingly, Fig.~\ref{fig:ranking_correlation} shows that the ranking among high-fitness individuals in early generations does not correlate with their final ranking, suggesting that evolutionary algorithms cannot make informed selections among good morphologies early in evolution, which poses a serious challenge for evolutionary brain-body co-optimization.

Together, these two findings suggest the existence of different learning speeds for different morphologies. While some morphologies achieve higher fitness with less controller optimization, other morphologies require extended controller optimization to achieve their maximum potential fitnesses. Similar findings demonstrating the existence of fast and slow learners are concurrently reported by~\citet{jeon2025convergent} using rigid robots, suggesting that this phenomenon is not limited to soft robots and can be more fundamental to the problem of brain-body co-optimization.

\begin{figure}[t]
    \centering
    \includegraphics[width=0.6\linewidth]{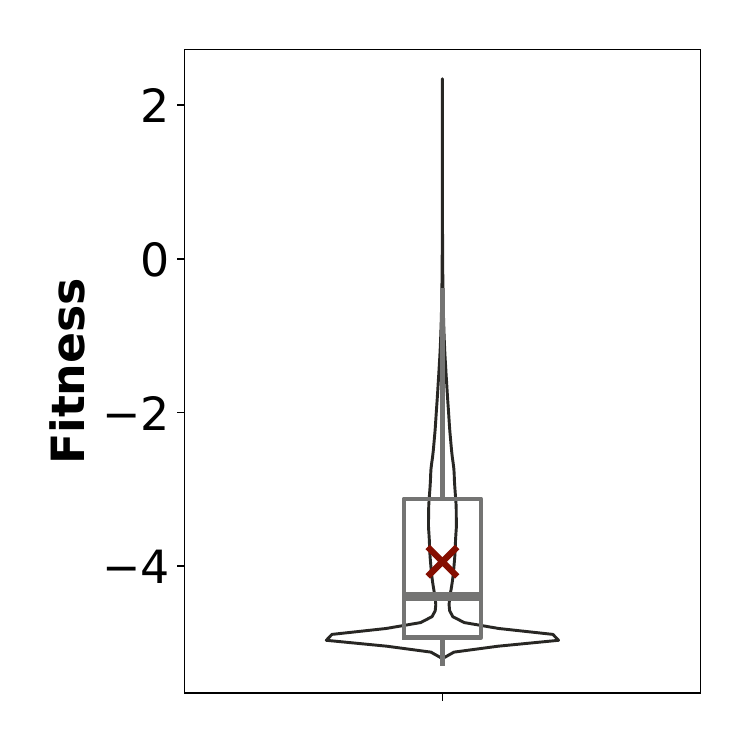}
    \caption{Distribution of fitness values in the 3-by-3 morphology space, based on 300 generations of evolutionary controller optimization. There are a large number of non-functional morphologies. The rest of the morphology space is more normally distributed in terms of fitness, with a heavy tail towards more fit morphologies. Although the design space is small, the distribution of fitness values does not suggest triviality or degeneracy. }
    \label{fig:full-fitness-dist}
\end{figure}

\paragraph{Is the 3-by-3 design space interesting?}
The 3-by-3 design space is small and it is natural to wonder whether it represents the challenges of designing morphologies. Here, we investigate how challenging the morphology-fitness landscape is. Fig.~\ref{fig:full-fitness-dist} shows the distribution of the true fitness values of each viable morphology in the morphology space of interest. The distribution shows a large group of robots that are centered at the bottom of the fitness range (below -4.4, approaching the -5 baseline for total immobility). These robots did not make any significant locomotion towards the goal during their lifetime, representing non-functional morphologies for the given problem. The remaining morphologies with fitness above -4.4 have a roughly normal distribution, centered around -3 (mean: -2.99, median: -3.12). Robots with a true fitness value of approximately -3 locomote toward the goal, exhibiting some level of competence. The distribution also shows a heavy tail toward higher fitness values, demonstrating the existence of a few exceptionally good morphologies, with the best morphology having a true fitness of 2.35. This data provides a validation point, demonstrating that the design space encompasses a well-balanced variety of morphologies in terms of fitness.

\begin{figure}
    \centering
    \includegraphics[width=0.8\linewidth]{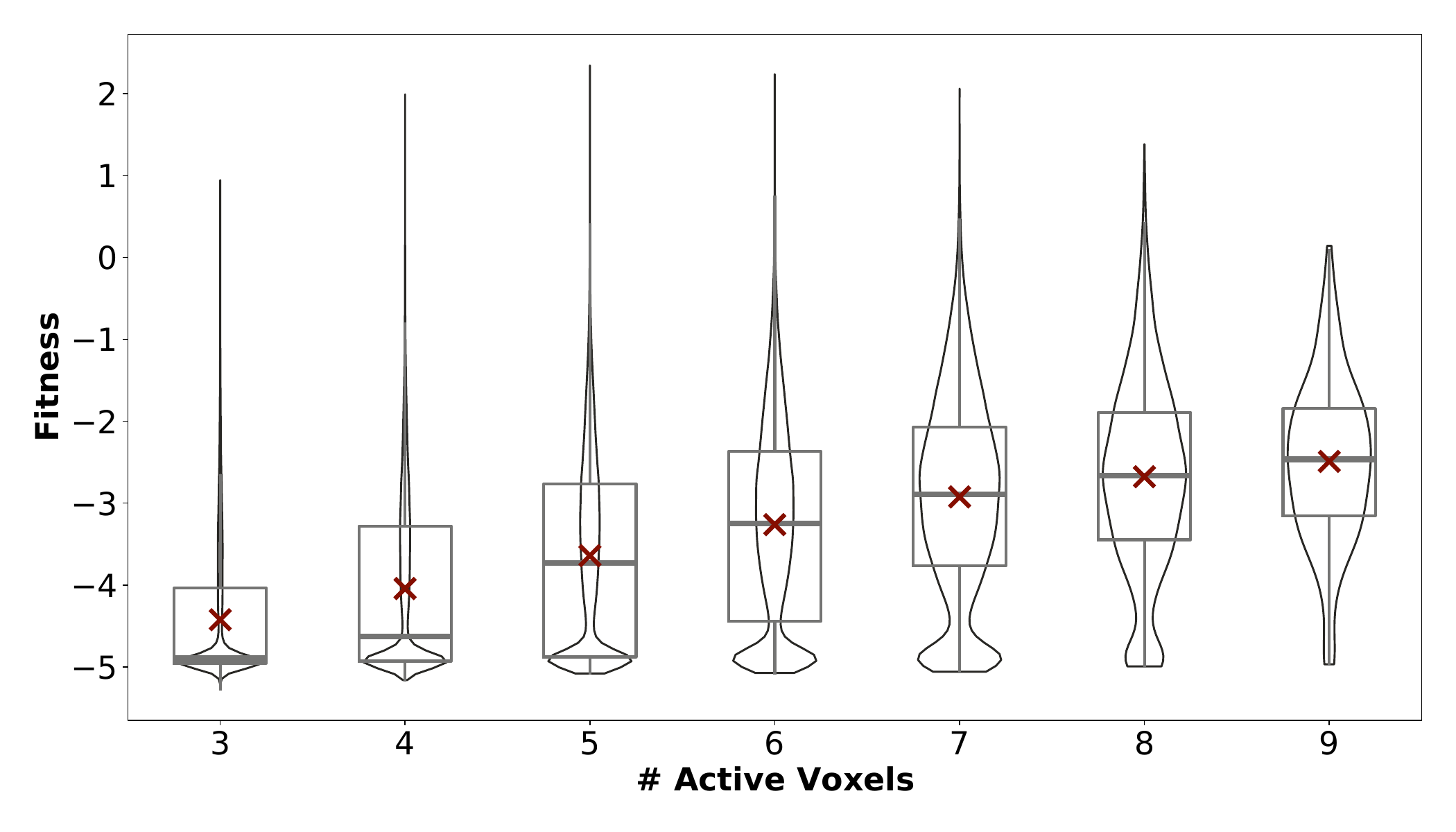} \\
    \includegraphics[width=0.8\linewidth]{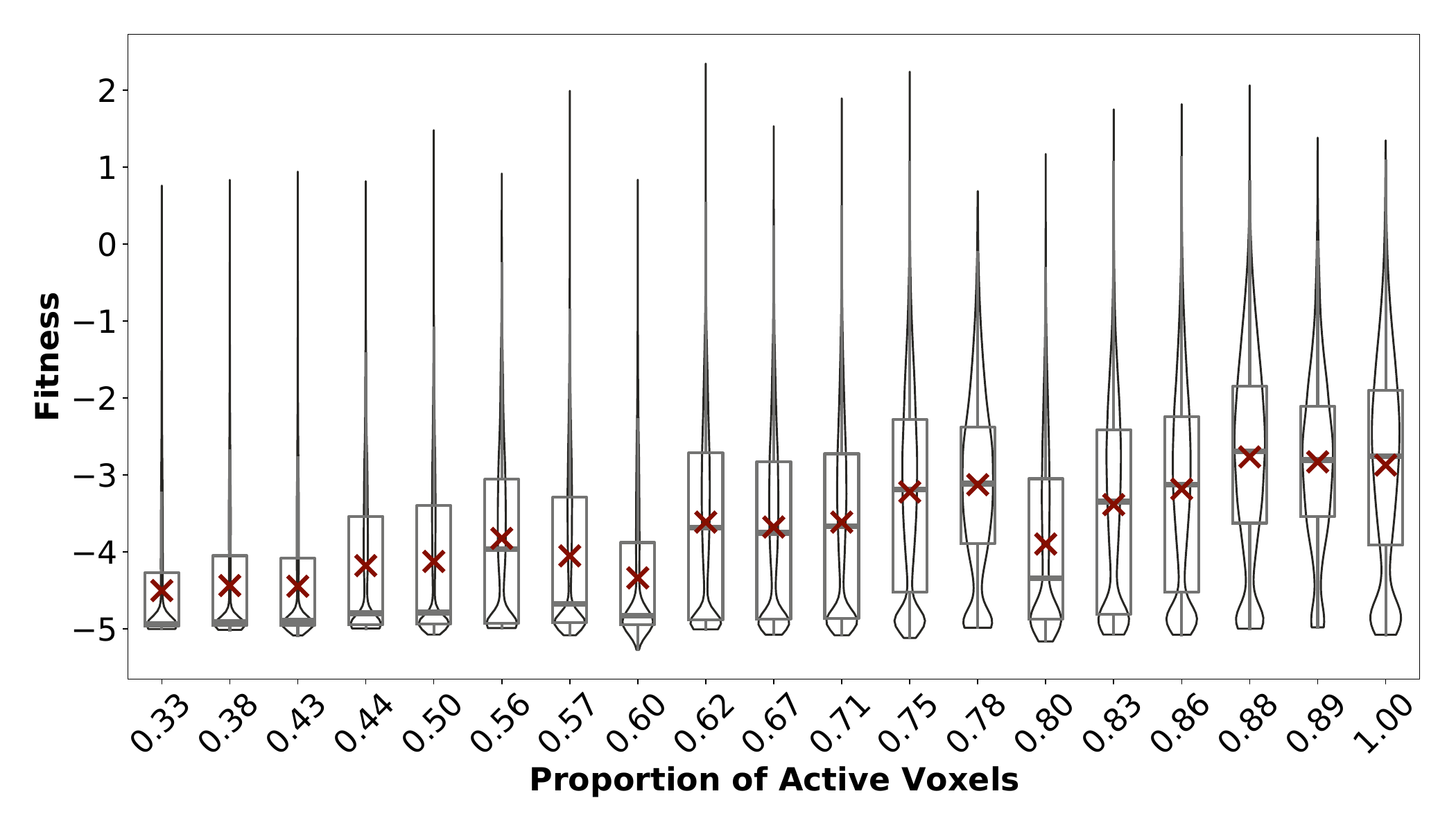}
    \caption{ The distribution of fitness values in the morphology-fitness landscape based on the number of active voxels in the robots' body (top) and the proportion of active voxels to all voxels (bottom). 
    Both plots show that the relation between the active voxels and fitness is not straightforward. The fitness of robots with the same number or proportion of active voxels can change greatly depending on how the voxels are placed in the robots' bodies. While the 3-by-3 design space is small, it captures the challenges of morphological design to an extent and provides a good test bed. 
    }
    \label{fig:body_analysis}
\end{figure}

Next, we investigate whether there is any trivial relation between morphology features and performance. Fig.~\ref{fig:body_analysis} provides data that demonstrate that the 3-by-3 design space is capable of providing a challenging design problem. The morphologies are grouped based on the number of active voxels (Fig.~\ref{fig:body_analysis} top) and their proportion to the whole body (Fig.~\ref{fig:body_analysis} bottom), and the distribution of the fitness values in each group is plotted. The data show that it is not the number of active voxels nor their proportion to the whole body that determines the locomotion performance, but rather their placements in the body that determine the fitness of morphologies, indicated by the wide distributions in each group. Moreover, the relation between active voxels and mean, median, or best performance in a group is not straightforward. We believe that this design space presents a challenging design problem.

Lastly, we discuss whether this design space is overly challenging, since even a single voxel change represents a large portion of the whole body of the robot and has potentially significant consequences. One can imagine that more complex robots, in the simplest case, bigger robots with more voxels, would not experience such consequences as the changes would be proportionally smaller. 
While this seems plausible, previous studies show that larger robots are similarly vulnerable to small changes that result in significant performance drops -- a phenomenon observed both generally \citep{mertan_modular_2023} and in comparative analyses \citep{medvet2022impact}.
In this sense, we argue that this small design space provides a suitable test bed to study brain-body co-optimization. 

\paragraph{How rugged is the morphology-fitness landscape?}
The morphology space can be considered as a network where each node corresponds to a distinct morphology, and the edges represent a single voxel change and connect two morphologies that are different at a single place in a 3-by-3 design grid. We define the local maxima based on the neighborhood in this network of morphologies. This conceptualization also provides a natural distance between morphologies. Working with this framework, we measure the average distances to certain points of interest to get a sense of the ruggedness of the space. The average distance to the nearest local maximum is 1.93, to the global maximum is 7.08, and to the near-optimal local maxima, those within the top 15\% of the fitness range, is 5.49.

\section{Brain-body Co-optimization} \hypertarget{sect:bb-coop}{}

Equipped with knowledge of estimated true fitness values of each morphology in the experimented design space, we run three different evolutionary brain-body co-optimization algorithms in the same morphology space to investigate how the evolutionary co-optimization unfolds and what parts of the search space are discovered. We examine three algorithms, age-fitness Pareto optimization (AFPO)~\citep{schmidt_age-fitness_2010}, MAP-Elites~\citep{mouret_illuminating_2015}, and morphological innovation protection (MIP)~\citep{cheney_scalable_2018}. 

AFPO~\citep{schmidt_age-fitness_2010} is designed to maintain a genetically diverse population of solutions. It selects based on performance and optimization time, where a solution can only outcompete another solution if it achieves higher fitness in fewer optimization steps. Combined with periodic injection of random solutions, this mechanism allows the algorithm to maintain some low performing solutions to increase genetic diversity in the population, helping the search process overcome local optima. MAP-Elites~\citep{mouret_illuminating_2015} is designed to maintain a phenotypically diverse population of solutions through explicit niching based on hand-designed phenotypic features of the solutions. Similarly to previous work where MAP-Elites was applied to the problem of brain-body co-optimization~\citep{nordmoen_map-elites_2021}, we choose the phenotypic features to be the number of active voxels in a morphology and the number of passive voxels. This allows the algorithm to maintain a population of solutions with diverse morphologies and potentially alleviate the premature convergence of morphology~\citep{joachimczak_artificial_2016,cheney_difficulty_2016}. Lastly, MIP~\citep{cheney_scalable_2018} is specifically designed to tackle fragile co-adaptation of morphology and control. It reduces the selection pressure on solutions with newly mutated bodies, preventing them from being prematurely eliminated from the population. 
The specific hyperparameters and simulation details for all three evolutionary algorithms are summarized in Table \ref{tab:ea_parameters}.

\begin{table}[htbp]
\centering
\caption{Summary of evolutionary algorithm parameters and simulation settings.}
\label{tab:ea_parameters}
\begin{tabular}{ll}
\hline
\textbf{Parameter} & \textbf{Value} \\
\hline
\textit{General Evolutionary Settings} & \\
Generations & 10,000 \\
Offspring Generation & 1 offspring per parent \\
Mutation Target & 50\% Brain / 50\% Body \\
Morphological Mutation & Single voxel change \\
Controller Mutation & Gaussian noise (mean 0, std 0.1) \\
Crossover & None \\
\hline
\textit{Algorithm-Specific Settings} & \\
MAP-Elites Archive Size & 28 \\
MAP-Elites Archive Dimensions & Number of Active voxels and Passive voxels \\
AFPO / MIP Population Size & 20 \\
AFPO / MIP Random Injection & 1 individual per generation \\
\hline
\textit{Simulation Details} & \\
Episode Length & Up to 500 time steps \\
Episode Termination Condition & Max steps or reaching the end of terrain\\
\hline
\end{tabular}
\end{table}

\subsection{Updating the Landscape}

Based on our prior findings showing that good morphologies require more controller optimization to estimate their maximum fitness well, each algorithm is run for 10,000 generations, giving them more chances to discover better controllers for morphologies that were not discovered in our initial morphology-fitness landscape creation. Indeed, for some morphologies, we found a better controller during the co-optimization process than we were able to find during the initial morphology-fitness landscape construction with 300 generations of controller optimization. We update the true fitness of 94,201 morphologies (20,249 of them found with AFPO~\citep{schmidt_age-fitness_2010}, 50,249 of them with MAP-Elites~\citep{mouret_illuminating_2015}, and 23,703 of them with MIP~\citep{cheney_scalable_2018}) with improved values and use this updated landscape for further analysis in the paper. While it increases the noise in the morphology-fitness landscape by introducing better estimations for some morphologies that are investigated further in brain-body co-optimization runs and leaving other morphologies' true fitness estimates unchanged, our further analysis uses these estimates as lower bounds and is not negatively affected by this noise.

\subsection{Analyzing Brain-body Co-optimization}

We use the updated morphology-fitness landscape data to analyze how three different evolutionary search algorithms unfold over evolutionary time. Specifically, we are interested in gaining insights into how optimization over morphologies takes place. First, we focus on the best solutions found at the end of each repetition (i.e., run champions) for each algorithm and their position in the search landscape. We start our inquiry by analyzing the performance of all algorithms.

\begin{figure}
    \centering
    \includegraphics[width=\linewidth]{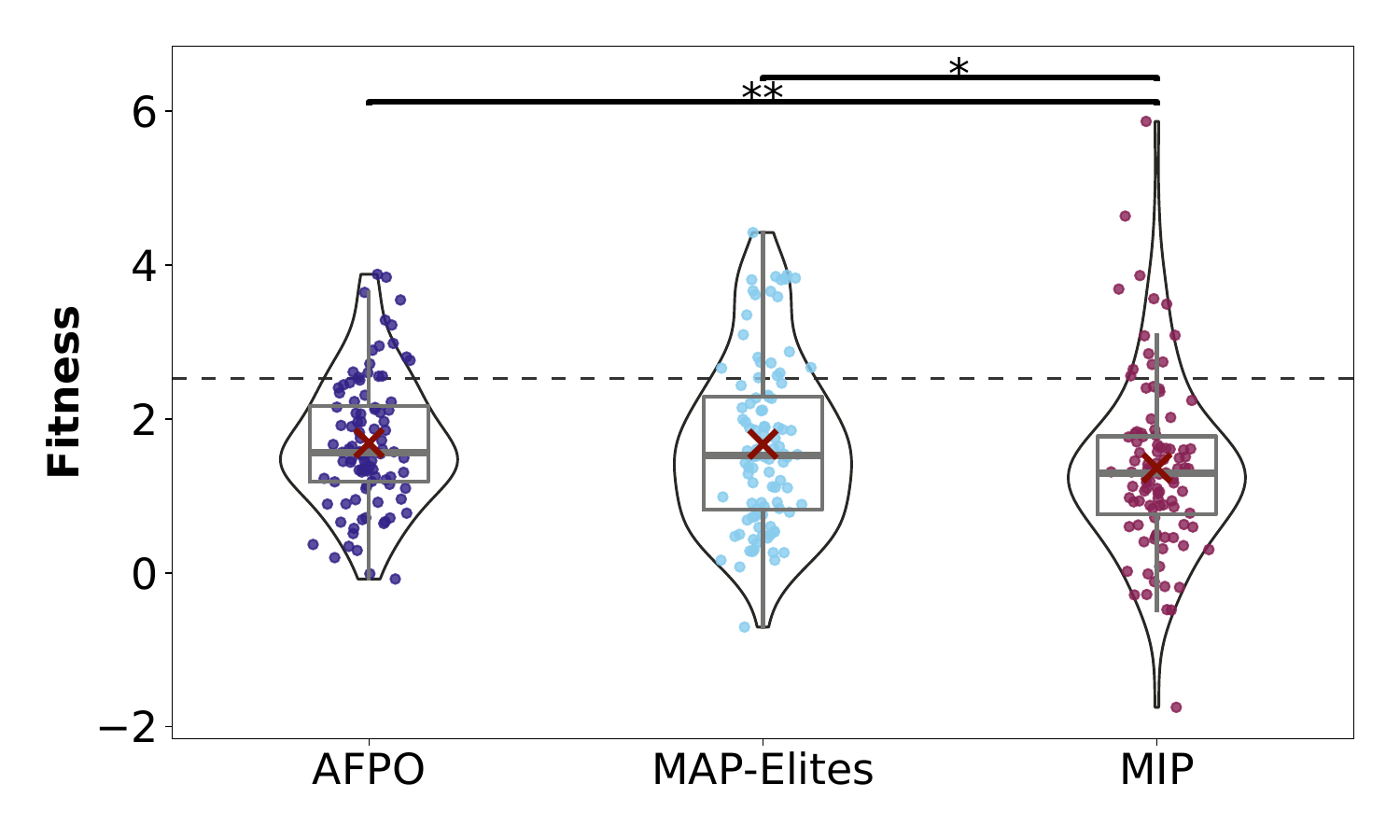}
    \caption{ 
     Fitness of the 100 run champions found during brain-body co-optimization with AFPO (left), MAP-Elites (middle), and MIP (right). The horizontal line shows the near-optimal threshold in the search space. In 100 independent repetitions, AFPO, MAP-Elites, and MIP are able to find a near-optimal solution 17, 22, and 13 times, respectively. Moreover, the distribution of run champions' fitness values for AFPO and MAP-Elites are statistically indistinguishable ($p >> 0.05$, Mann–Whitney U test~\citep{mann1947test}), indicating that the extra morphological diversity maintained with MAP-Elites does not lead to good stepping stones for a better run champion. MIP, on the other hand, performs slightly worse (*: $p < 0.05$, **: $p < 0.01$), however finds the best run champion across all of the runs. These results suggest that experimented algorithms have different optimization dynamics, yet all of them fail to find high-performing solutions in majority of the independent runs.
    }
    \label{fig:rc-coop-vs-updated}
\end{figure}

\paragraph{Do evolutionary brain-body co-optimization algorithms find near-optimal solutions?}
Fig.~\ref{fig:rc-coop-vs-updated} shows the fitness distribution of run champions found during evolutionary brain-body co-optimization with the experimented algorithms. The horizontal line shows the threshold for near-optimality, which is generously defined as the top $30\%$ of the fitness range -- 
\begin{equation}
  \begin{aligned}
      &\text{Near-optimality Threshold}^{(=2.53)}  = \\
      &(\text{Global Max}^{(=5.87)} - \text{Global Min}^{(=-5.27)}) * 0.7 \\
      & + \text{Global Min}^{(=-5.27)}.
  \end{aligned}
\end{equation}
There are 308 morphologies that achieve near-optimal performance. The AFPO algorithm is able to find a near-optimal morphology 17 times out of 100 independent trials, while the MAP-Elites algorithm finds 22 times and MIP finds 13 times. \textit{All of the experimented algorithms fail to find near-optimal solutions in the majority of the independent runs, demonstrating the difficulty of brain-body co-optimization}. Additionally, the distribution of run champions' fitness values for AFPO and MAP-Elites are statistically indistinguishable, suggesting that the explicit niche construction to maintain morphological diversity does not lead to a better global search in morphology space.\footnote{We emphasize that our implementation of MIP differs from the original version by~\citet{cheney_scalable_2018}. In their work, a threshold on morphological change is used to decide whether offspring receive reduced selection pressure. In our setting, however, body mutations consist of only a single voxel modification, so every offspring with a newly mutated body is given reduced selection pressure. This distinction may account for the discrepancy between our results and those reported in the original study.} MIP, however, finds the best solution across all independent runs, indicating that these algorithms have distinct optimization dynamics. 

Next, we investigate the reasons why evolutionary search algorithms show poor performance in the brain-body co-optimization problem. A natural thing to investigate is to see whether these algorithms get stuck in local maxima in the morphology space. 

\begin{figure}[t]
    \centering
    \includegraphics[width=\linewidth]{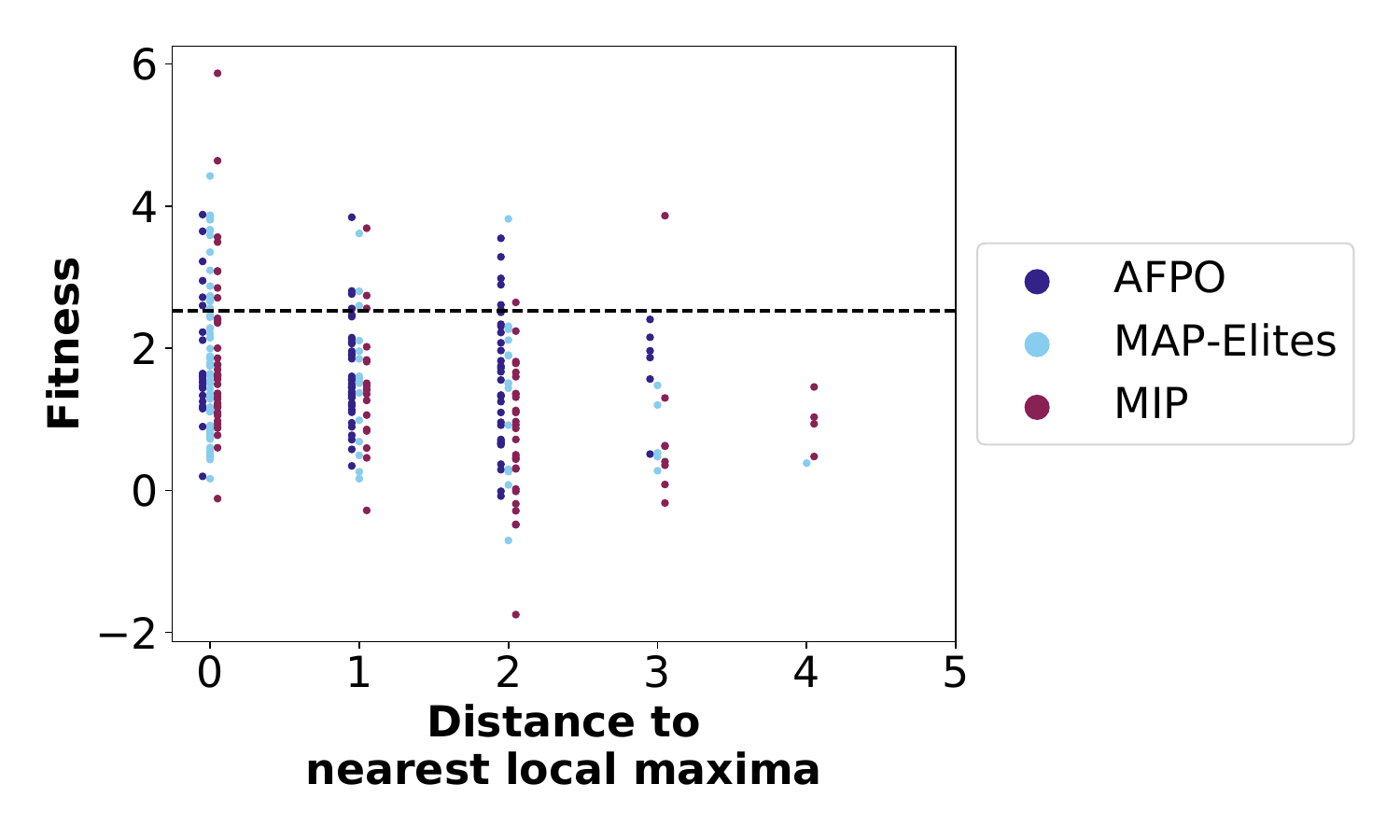}
    \caption{ 
    The fitness values of run champions (x-axis) are plotted against the distance to the nearest local maximum in the morphology space that they are in the basin of (y-axis). Run champions of different treatments are plotted with a small vertical separation for visual clarity. 
    Run champions with a distance of zero indicate local maxima. There are 22 run champions that are local maxima for AFPO, 62 for MAP-Elites, and 43 for MIP. MAP-Elites more consistently exploits the local neighborhood of solutions in the morphology space.
    Strikingly, however, 38 out of 100 independent trials for MAP-Elites get stuck at a point in the morphology space that is not even a local maximum (This number is 78 for AFPO and 57 for MIP). This indicates that all experimented algorithms struggle to track the fitness gradient in the morphology-fitness landscape.
    }
    \label{fig:rc-coop-vs-updated-lm-analysis}
\end{figure}

\paragraph{Are run champions found in evolutionary brain-body co-optimization local optima in the morphology space (i.e., does search get stuck in a local optimum)?}
Fig.~\ref{fig:rc-coop-vs-updated-lm-analysis} plots the fitness of the run champions against their distance to the nearest local maximum that they are in the basin of, in the morphology space. Run champions with a distance of zero are local maxima in the morphology space, indicating that the search can exploit the local structure in the morphology-fitness landscape. The number of run champions that are local maxima is 22 for AFPO, 62 for MAP-Elites, and 43 for MIP (out of 100). MAP-Elites finds the local optima in the morphology-fitness landscape more consistently than the other two algorithms, successfully exploiting the local structure. The better local optimization observed in the MAP-Elites could be due to the explicitly maintained morphological diversity. 

However, even in the case of MAP-Elites, 38 of 100 independent repetitions get stuck at a point in the morphology space that is not even a local maximum. \textit{Strikingly, experimented algorithms fail to discover morphologies that have higher estimated true fitness, despite scenarios where just one voxel alteration is sufficient.} Note that our claim that these algorithms get stuck in points that are not even local optima is only true in the morphology space. In the combined solution space of controller-morphology pairs, algorithms are likely to be stuck at local maxima. 

\begin{figure}[t]
    \centering
    \includegraphics[width=\linewidth]{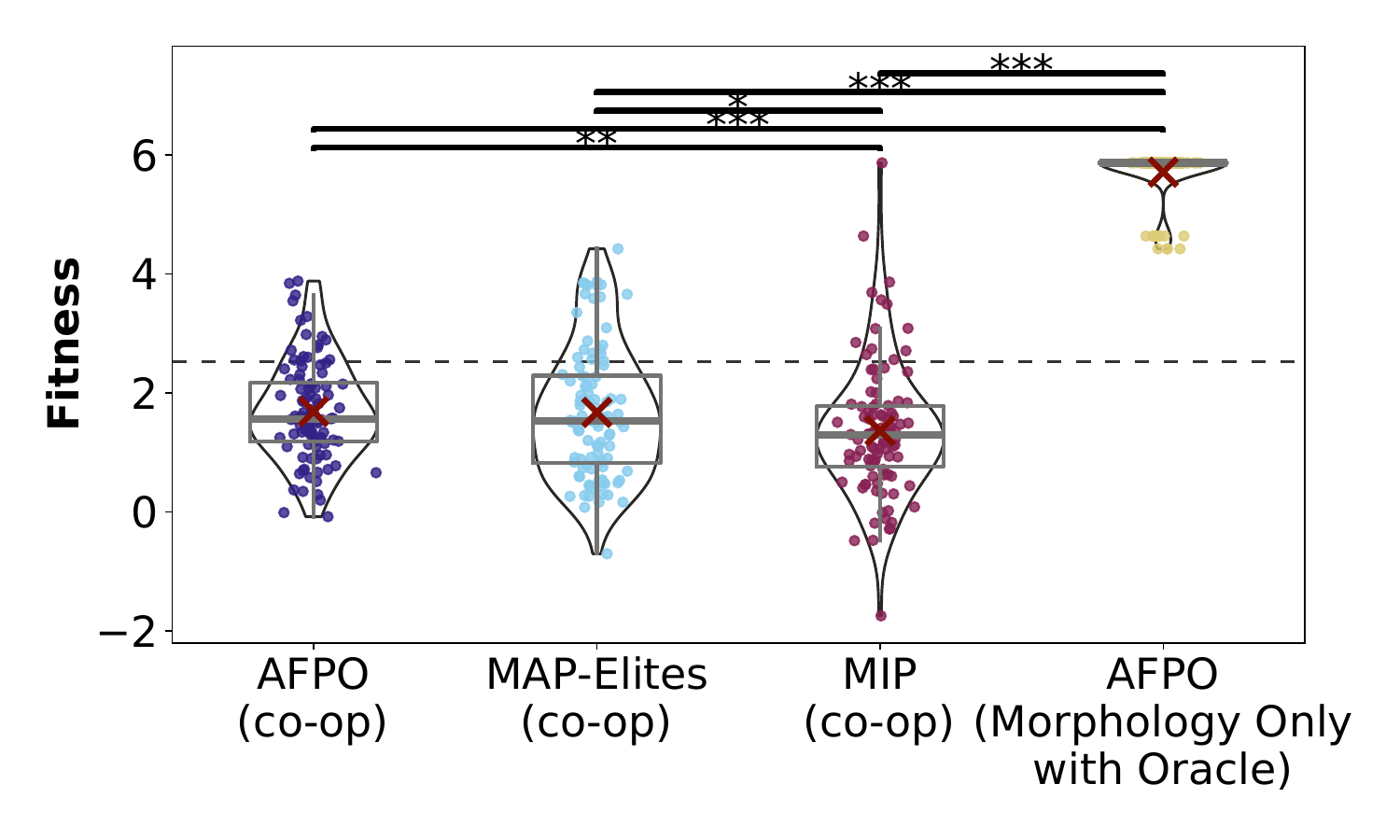}
    \caption{ 
    Distribution of run champions' fitness for brain-body co-optimization with AFPO and MAP-Elites, compared to morphology-only evolution with AFPO. 
    The dotted horizontal line shows the near-optimality threshold. 
    The solid horizontal lines show statistically significantly different distributions  (*: $p < 0.05$, **: $p < 0.01$, ***: $p < 0.001$, Mann–Whitney U test~\citep{mann1947test}).
    Evolving morphologies with access to their true fitness values can achieve statistically significantly better results, with less computational resources. While true fitness estimates are inaccessible in practice, this experiment demonstrates that the fitness gradients in the morphology-fitness landscape are easy to track in the absence of simultaneous controller optimization.
    }
    \label{fig:coop-rc-vs-ms-rc}
\end{figure}

To support this conclusion, we run 100 repetitions of morphology-only evolution, where the fitness of each morphology is given as its estimated true fitness values. We employ the AFPO algorithm with considerably reduced computational resources than those used in brain-body co-optimization experiments. This is achieved by employing a smaller population size of 10 and conducting 1,000 generations of evolution, in contrast to using 20 and 10,000, respectively\footnote{In co-optimization experiments, every individual in the population generates an offspring through either brain or body mutation. By reducing the population size by half, we efficiently manage the computational resources allocated in exploring the morphology space each generation.}. Fig.~\ref{fig:coop-rc-vs-ms-rc} plots the distribution of run champions' fitness values for brain-body co-optimization with the experimented algorithms, as well as for morphology-only evolution with AFPO. Morphology-only evolution statistically significantly outperforms brain-body co-optimization and consistently finds near-optimal morphologies, despite using fewer computational resources. This experiment demonstrates that evolutionary search algorithms can track fitness gradients in the morphology-fitness landscape efficiently; however, simultaneously optimizing control significantly hinders the search over the morphology-fitness landscape. This finding aligns with previous research that indicates that evolutionary algorithms perform better when controllers are fixed, rather than evolving controllers and morphologies simultaneously~\citep{mertan_investigating_2024}.

\begin{figure}[t]
    \centering
    \includegraphics[width=\linewidth]{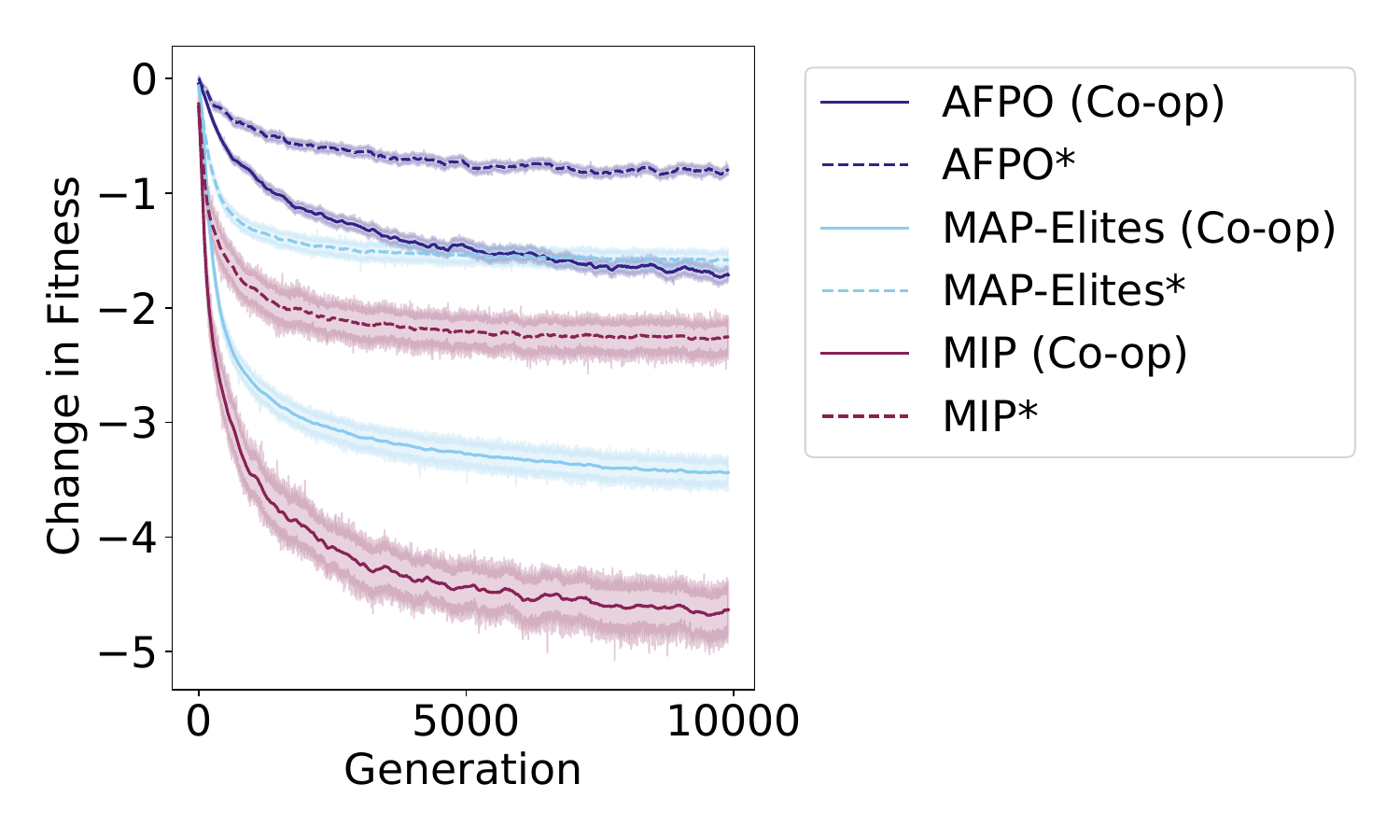}
    \caption{ 
    The average change in fitness, when an offspring is created through body mutations, is plotted against evolutionary time. Solid lines show the average across 100 independent repetitions and the shaded areas show the 95\% boosted confidence intervals. The data is smoothed temporally by applying moving average with a window size of 100 to make trend more apparent.
    While AFPO, MAP-Elites, and MIP plot the change in fitness observed during evolution; AFPO*, MAP-Elites*, and MIP* plot the change in estimated true fitness. 
    For all experimented algorithms, fitness drops during co-optimization are significantly worse than the estimated true fitness difference between morphologies. 
    This shows that algorithms are likely to undervalue new individuals with newly mutated morphologies due to significant negative transfer of controllers. Interestingly, the extent of this undervaluation is worse for MAP-Elites.
    }
    \label{fig:coop-fitness-change}
\end{figure}

\paragraph{How does simultaneously optimizing controllers negatively affect the co-optimization process?} 
Previous work~\citep{joachimczak_artificial_2016,cheney_difficulty_2016,cheney_scalable_2018,mertan_modular_2023,mertan_investigating_2024} points out an important phenomenon, fragile co-adaptation: controllers that are optimized for particular morphologies do not transfer well to others. Negative transfer results in low-performing solutions whenever the morphology of a solution is changed, i.e., whenever a new offspring is created with a body mutation. It is hypothesized that the negative transfer results in the elimination of promising morphologies, as evolutionary search algorithms tend to discard such low-performing solutions, and the search over the morphology space gets stuck in non local-maxima. Although previous work has shown the negative transfer of controllers to different morphologies empirically~\citep{mertan_modular_2023} and discussed its results~\citep{mertan_investigating_2024,mertan_controller_2025}, here having access to true fitness values of all morphologies in the search space allows us to concretely demonstrate the elimination of promising morphologies throughout evolution.

\begin{figure}[t]
    \centering
    \includegraphics[width=\linewidth]{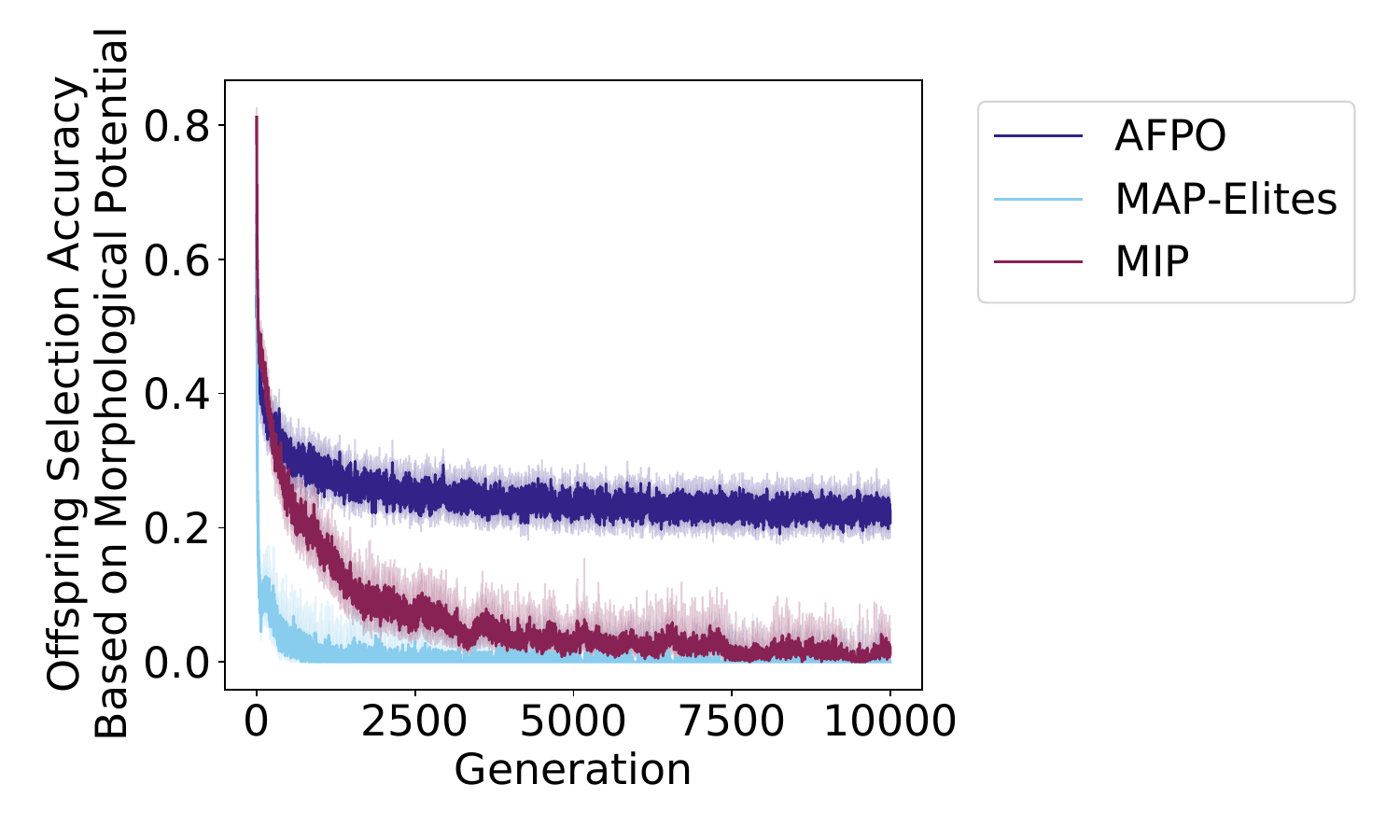}
    \caption{ 
    The accuracy of selection based on morphological potential (i.e., estimated true fitness) is plotted against evolutionary time. We retrospectively analyze the evolutionary brain-body co-optimization runs, and perform counterfactual selections using the estimated true fitness values at every generation. Higher accuracy means the offspring with higher estimated maximum fitness are selected, lower accuracy means the offspring with higher estimated maximum fitness are eliminated in favor of solutions with lower estimated maximum fitness.
    The results show that all of the experimented algorithms consistently eliminate offspring that would have been selected if their true fitness were known. This demonstrates the consequences of the undervaluation of individuals with newly mutated morphologies -- solutions with higher morphological potential are being eliminated.
    }
    \label{fig:coop-elimination}
\end{figure}

Fig.~\ref{fig:coop-fitness-change} plots the average change in fitness when an offspring is created through body mutations during brain-body co-optimization with the experimented algorithms. The average change in estimated true fitness is also plotted for comparison. The results illustrate that the fitness drop observed in applying body mutations during brain-body co-optimization is worse compared to the fitness drop in estimated true fitness. Algorithms undervalue the fitness of individuals with newly mutated morphologies. Fig.~\ref{fig:coop-elimination} quantifies the outcome of this undervaluation, in which selections happened during evolution are compared to \textit{counterfactual selections}, i.e., selection that would have happened if the algorithm had access to estimated true fitness values.
It shows that all the experimented algorithms eliminate promising morphologies at high rates throughout evolution, preventing algorithms from tracking fitness gradients in the morphology-fitness landscape effectively. \textit{In essence, when optimizing controllers simultaneously, it results in fitness valleys between adjacent morphologies, complicating the task of search algorithms to identify nearby high-performing morphologies and track the fitness gradient within the morphology-fitness landscape.}

Interestingly, AFPO, compared to the other two algorithms, is slightly better at not undervaluing solutions with newly mutated morphologies (Fig.~\ref{fig:coop-fitness-change}), and it is better at selecting offspring based on morphological potential (Fig.~\ref{fig:coop-elimination}). However, this does not translate to better run champions (Fig.~\ref{fig:rc-coop-vs-updated}) or better local optimization (Fig.~\ref{fig:rc-coop-vs-updated-lm-analysis}). The connection between the different mechanisms utilized by the experimented algorithms and their efficacy in locating globally and locally optimal solutions appears to be complex and warrants further exploration.

\section{To Co-optimize or Not to Co-optimize?} \hypertarget{sect:goal-switch}{}

We have identified the challenge of evolutionary brain-body co-optimization as the inability to assess the true fitness of morphologies. One simple approach to tackle this is to decouple the optimization of brain and body: use a bi-level optimization scheme where the outer loop explores different morphologies, and for each sampled morphology, the inner loop fully optimizes its controller to better estimate its performance, effectively eliminating the issue. Although computationally demanding, this method has been used to address the problem of brain-body optimization~\citep{gupta_embodied_2021,strgar_evolution_2024}. Note that in this approach, each controller is trained for a particular morphology. This stands in sharp contrast to brain-body co-optimization, where both components of the solution are continuously evolving. 

\begin{figure}
    \centering
    \includegraphics[width=\linewidth]{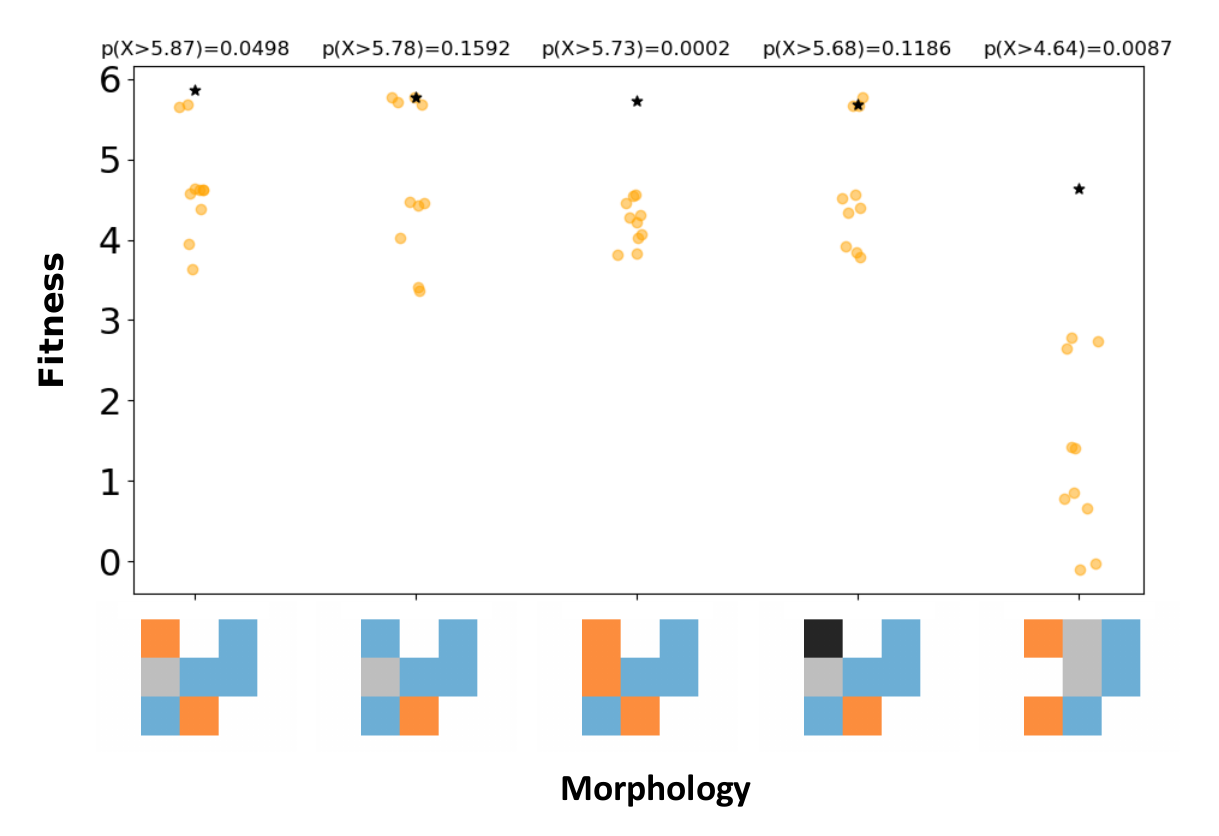}
    \caption{ 
    Comparison between the performances obtained via evolutionary brain-body co-optimization (black stars) and those reached by optimizing only the controllers while keeping the morphologies fixed (yellow dots), shown for the five best-performing morphologies (shown in the bottom x-axis) according to our true fitness estimates. The top x-axis shows the probability of achieving higher performance with control-only optimization, when we fit a t distribution to the data of the control only optimization. In two out of five cases we tested, optimizing a controller from scratch for the tested morphology is unlikely ($< 0.01$) to produce a controller as good as one that is found through evolutionary brain-body co-optimization. This suggests that these control problems have a deceptive optimization landscape where optimizing the controller directly for the target morphology gets stuck in a local optimum.
    }
    \label{fig:coop-vs-control}
\end{figure}

Here we investigate this difference to see if co-optimization has any benefits, other than the supposed reduction in computational cost. Inspired by recent work of~\citet{templier2025time} showing how changing actuator torques throughout optimization (mimicking biological developmental changes in natural organisms) outperforms training directly with the final model for legged locomotion tasks, we hypothesize that \textit{evolutionary brain-body co-optimization can create a path in the solution space that mimics ontogenetic development of morphology, and it results in higher performance compared to optimizing a controller for the final form of the morphology.} 

To test this hypothesis, we focus on the top five highest performing morphologies in the design space (shown in Fig.~\ref{fig:coop-vs-control}). Interestingly, the best controller for each one of these morphologies was found during evolutionary co-optimization, specifically with the MIP algorithm. This performance is marked with a black star in Fig.~\ref{fig:coop-vs-control}. We compare this performance obtained through co-optimization (co-op treatment) to the case where the morphology is kept fixed and only the controller is optimized (control-only treatment). For a fair comparison in terms of number of simulation steps used to search the space of controllers, control-only treatment runs the same number of generations as the co-op treatment (10,000 generations), but half the population size (20 for co-op treatment, 10 for control-only treatment), as the co-op treatment uses half of its computational resources searching the space of morphologies. We run 10 independent repetitions of the control-only treatment with different random seeds and fit a t distribution to this data to test the probability of control-only treatment finding a controller as good as co-op treatment.

\begin{figure}
    \centering
    \includegraphics[width=\linewidth]{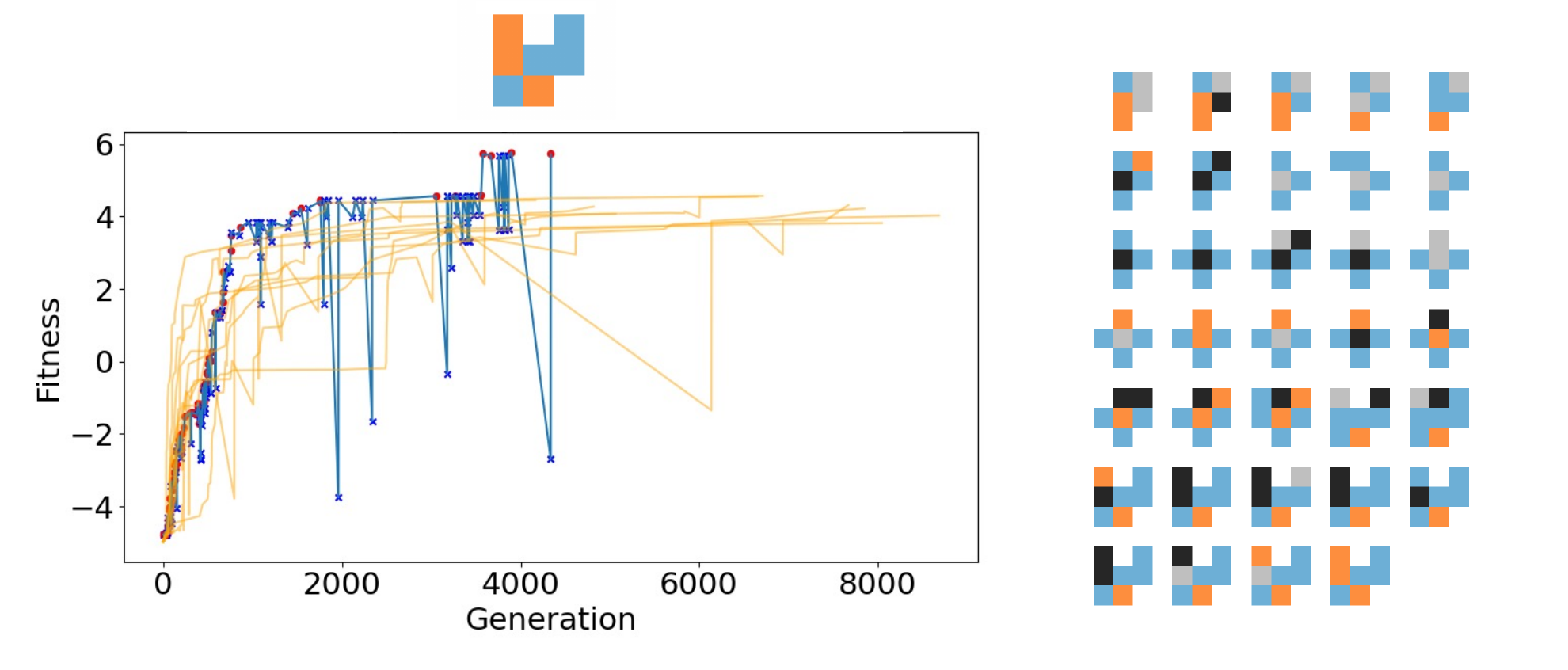} \\
    \includegraphics[width=\linewidth]{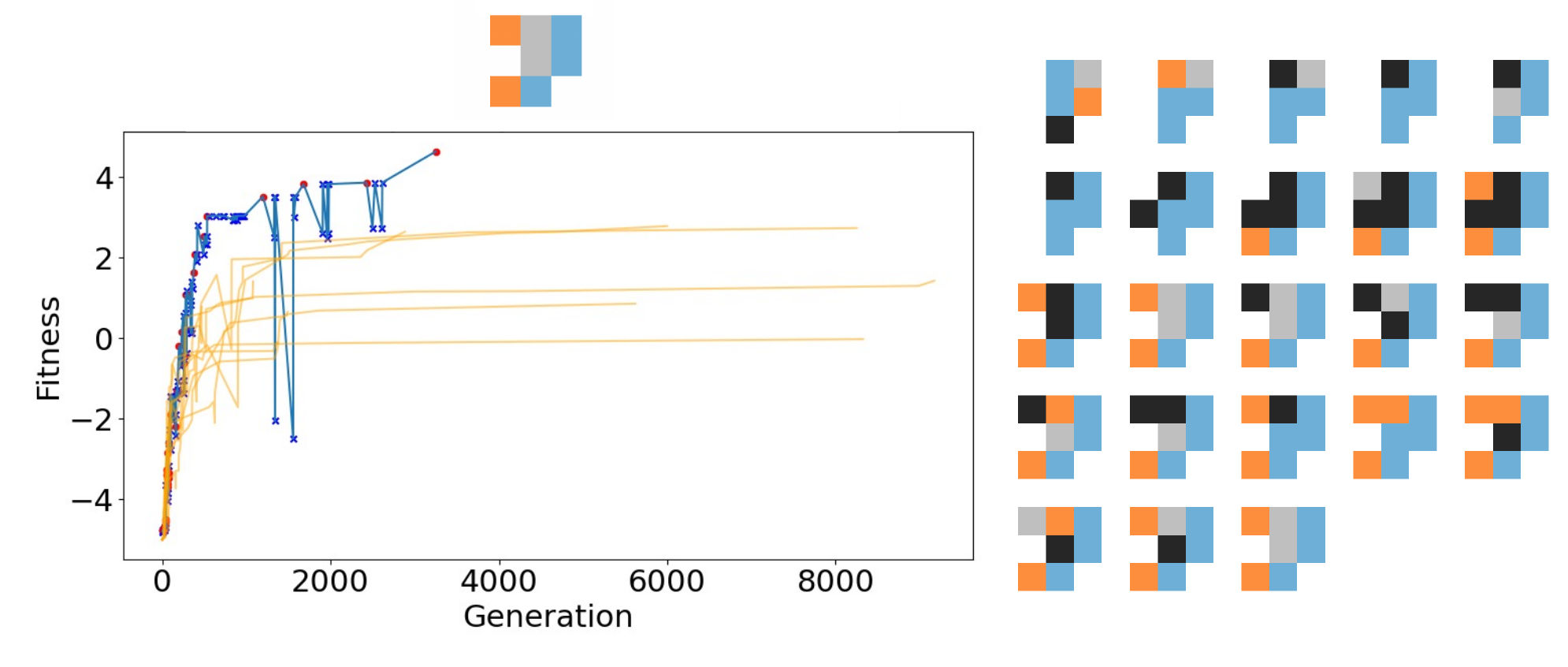}
    \caption{ 
    A detailed examination of how optimization unfolds for two of the five morphologies (shown in the top x-axes) in which control-only optimization fails to reach the performance level achieved by evolutionary brain–body co-optimization. 
    The blue lines trace the fitness of the lineage of the best-performing solutions discovered for these two morphologies during brain–body co-optimization, plotted against the number of optimization steps. The blue crosses mark the ancestors created through body mutations, and the red dots mark the ancestors created through brain mutations. 
    The yellow lines trace the fitness of the run-champions' lineage from each of the 10 control-only optimization runs. 
    During evolutionary brain–body co-optimization, controllers are evaluated and selected based on how well they perform across a wide range of morphologies. For the lineages of interest, these morphologies are shown on the right, arranged in reading order, with the top-left morphology representing the initial form and the bottom-right morphology representing the final form. We hypothesize that this creates useful goal-switching that helps the optimization process avoid local optima and find higher-performing controllers for the final morphologies in fewer optimization steps compared to optimizing the controllers only for the final morphologies.
    }
    \label{fig:coop-vs-control-detailed}
\end{figure}

Fig.~\ref{fig:coop-vs-control} shows that in two out of five cases, control-only treatment is unlikely to achieve the same level of performance as co-op treatment. This suggests that the optimization landscapes of the controllers for these two morphologies are deceptive~\citep{lehman_abandoning_2011,lehman2011novelty} -- optimizing the controller directly for these morphologies gets stuck in local optima. Fig.~\ref{fig:coop-vs-control-detailed} provides a detailed look at how optimization unfolds for these cases.
The blue lines trace the fitness of the lineage of the best-performing solutions discovered by co-op treatment, plotted against the number of optimization steps. The blue crosses mark the ancestors created by body mutations, and the red dots mark the ancestors created by brain mutations. The yellow lines trace the fitness of the run-champions' lineage from each of the 10 control-only optimization runs.
Control-only treatment gets stuck in a local optimum and fails to make progress in independent runs. On the other hand, during evolutionary brain–body co-optimization, controllers are evaluated and selected based on how well they perform on a trajectory of morphologies, which are shown on the right, arranged in reading order. We believe that this creates useful goal-switching~\citep{nguyen2015innovation} that helps the optimization process avoid local optima and find controllers that perform better for the final morphologies in fewer optimization steps compared to optimizing controllers only for the final morphologies.

This finding has important implications. It suggests that the common practice of hand-designing a robot morphology and training a controller from scratch for that morphology would result in a suboptimal performance, not only because it is hard to design a morphology by hand, but also because training a controller for a fixed morphology results in a suboptimal controller. It puts into question whether a bi-level scheme for brain-body optimization is effective as well, since it decouples the optimization of morphology and control. 

It also serves as an important demonstration of what embodiment means for intelligence. Previous work demonstrates that, by virtue of being embodied, an agent can exploit the morphology of its body to solve problems that would otherwise require complex solutions~\citep{ziemke2002neuromodulation}. Moreover, the form of the body can determine its function~\citep{maris1996exploiting, braitenberg1986vehicles}, and highly optimized morphologies can effectively exploit the passive dynamics of the body~\citep{mcgeer1990passive,beal2006passive}. More recent results reveal that the body’s compliance can itself perform non-linear computation, morphological computation~\citep{hauser2011towards} -- offloading a substantial portion of the computational burden from the brain to the body. The specific features of embodiment, in turn, influence how we perceive, act, and think~\citep{pfeifer_how_2007,lakoff2008metaphors}, and even produce complex adaptive behavior~\citep{mertan2024no,mertan2025morphological}. In this work, we provide preliminary evidence that the ontogenetic development of the body can create useful goal-switching for the brain, improving its learning dynamics. 

We acknowledge that the results presented here do not conclusively prove the hypothesis that we put forward. One can argue that the comparison of co-op treatment and control-only treatment is not fair, as we had to run three different evolutionary co-optimization algorithms, 100 times each, with twice the computational resources per run compared to control-only treatment to find cases to support our hypothesis. However, fair comparison is challenging as control-only treatments require the knowledge of which morphologies to test a priori, which is not accessible in practice, especially considering the unintuitive designs of the top five highest performing morphologies based on our estimations. 

Instead, we state that the results suggest a potential benefit for co-optimization, and we encourage further research on this phenomenon. The prevalence of this phenomenon, as well as the question of whether optimization focused solely on control could reach comparable performance given greater computational resources, remains to be examined. Importantly, investigations are needed into methods for enhancing brain–body co-optimization algorithms, with the aim of promoting more favorable developmental trajectories.

\section{Discussion} \hypertarget{sect:discussion}{}

Considering the finding that it is not possible to rank high-performing individuals early in evolution (as it requires more controller optimization to estimate their fitness well, as shown in Fig.\ref{fig:ranking_correlation}), and persistent issue of eliminating morphologies with higher true fitness throughout evolution (Fig.~\ref{fig:coop-elimination}), we hypothesize that evolutionary co-optimization algorithms lead to dynamics that give rise to something akin to a type of first-mover advantage~\citep{lieberman_first-mover_1988} -- customers (the algorithms) are less likely to switch to a new product (select a morphology with higher true fitness for survival) when there is a cost associated with adapting to the new product (controller adaptation to the new morphology to accurately estimate the true fitness of the new morphology). That is to say, morphologies that are selected in an uninformed way early in evolution accrue more controller optimization and start to outcompete late-rival morphologies with higher true fitness but lower initial fitness due to negative transfer.

Our findings also ground the important trends in the field. We identify the problem of brain-body co-optimization as the inability to estimate the true fitness of morphologies with strong evidence. This explains why ``if it evolves, it needs to learn''~\citep{eiben_if_2020} from a practical point of view -- controller learning remedies negative transfer of the controllers and improves fitness estimates of the morphologies. The success of recent methods can be understood from this perspective, such as learning with adaptive length~\citep{cheney_scalable_2018,le_goff_improving_2024}, lifetime learning~\citep{ferigo2025totipotent} (which supports a Lamarckian evolutionary scheme that might offer additional benefits~\citep{luo2023lamarck}), or even bi-level optimization where the outer loop optimizes the morphology and the inner loop fully optimizes a controller for each morphology~\citep{gupta_embodied_2021,strgar_evolution_2024} (which can leverage advances in sample efficient learning~\citep{yu2018towards,xu2022accelerated}). In the latter case, however, fully decoupling optimization of morphology and control might result in suboptimal performance, as suggested by our results.

Another line of work tackles the brain-body co-optimization by improving the transfer of controllers to unseen morphologies. Decentralized control~\citep{pigozzi_evolving_2022,mertan_modular_2023}, complex neural network architectures~\citep{gupta_metamorph_2022, wang_nervenet_2018}, and multi-morphology controllers~\citep{mertan2024towards, mertan_controller_2025, jeon2025convergent} are investigated to make brains more robust to morphological changes. These approaches also have the advantage of synergizing well with the work that introduces algorithmic improvements.

\citet{zhao_robogrammar_2020,wang2023preco,song2024morphvae} learn generative models that are optimized to produce better morphologies. These approaches try to avoid the issue of negative transfer-related undervaluation. Alternatively, \citet{matthews2023efficient} implement a differentiable simulator to achieve gradient-based design optimization, enabling better gradient tracking in the morphology-fitness landscape. More research is required to see how these approaches compare to others in terms of performance and to see if they can scale up to more complex problems.

In future work, we hope to use the morphology-fitness landscape we produce in this work to investigate the effectiveness of current approaches. This will also involve considering experimental design decisions, including alternative training schedules, different hyperparameters, and the selection of encodings, as previous work has highlighted its relevance in both single-level \citep{ferigo2022entanglement} and bi-level \citep{pigozzi2023morphology} optimization. Moreover, we plan to extend this analysis in future work to more challenging design spaces, potentially utilizing different simulation engines, to test the generality of the insights derived in this work.

\section{Conclusion}
We provide a deeper understanding of the brain-body co-optimization problem by exhaustively mapping a morphology-fitness landscape. We hope this data is useful to the community and foster more research in this domain. To show its usefulness, we conduct an analysis of three evolutionary algorithms, namely age-fitness Pareto optimization (AFPO)~\citep{schmidt_age-fitness_2010}, MAP-Elites~\citep{mouret_illuminating_2015}, and morphological innovation protection (MIP)~\citep{cheney_scalable_2018}, applied to the problem of brain-body co-optimization in the same morphology-fitness landscape. Equipped with the knowledge of the complete morphology-fitness landscape, we show that experimented evolutionary algorithms fail to discover near-optimal solutions and, at times, get stuck in morphologies that are not even local maxima. In essence, simultaneous optimization of controllers creates ``fitness valleys'' between morphologies and creates dynamics similar to first-mover advantages, hindering the evolutionary optimization process. On the other hand, co-optimization of brain and body also has the potential of finding higher-performing solutions through goal-switching. Our findings help to understand trends in the field and provide valuable insights for future work.

\section{Acknowledgements}
This material is based upon work supported by the National Science Foundation under Grant No. 2239691 and 2218063.  Thanks to the Vermont Advanced Computing Center for providing computational resources.

\printbibliography

@inproceedings{sims_evolving_1994,
    address = {Not Known},
    title = {Evolving virtual creatures},
    isbn = {978-0-89791-667-7},
    url = {http://portal.acm.org/citation.cfm?doid=192161.192167},
    doi = {10.1145/192161.192167},
    language = {en},
    urldate = {2022-03-04},
    booktitle = {Proceedings of the 21st annual conference on {Computer} graphics and interactive techniques  - {SIGGRAPH} '94},
    publisher = {ACM Press},
    author = {Sims, Karl},
    year = {1994},
    pages = {15--22},
}

@inproceedings{mertan_investigating_2024,
    address = {Cham},
    title = {Investigating {Premature} {Convergence} in {Co}-optimization of {Morphology} and {Control} in {Evolved} {Virtual} {Soft} {Robots}},
    isbn = {978-3-031-56957-9},
    booktitle = {Genetic {Programming}},
    publisher = {Springer Nature Switzerland},
    author = {Mertan, Alican and Cheney, Nick},
    editor = {Giacobini, Mario and Xue, Bing and Manzoni, Luca},
    year = {2024},
    pages = {38--55},
}

@inproceedings{mertan_modular_2023,
    address = {Lisbon Portugal},
    title = {Modular {Controllers} {Facilitate} the {Co}-{Optimization} of {Morphology} and {Control} in {Soft} {Robots}},
    isbn = {9798400701191},
    url = {https://dl.acm.org/doi/10.1145/3583131.3590416},
    doi = {10.1145/3583131.3590416},
    language = {en},
    urldate = {2023-10-20},
    booktitle = {Proceedings of the {Genetic} and {Evolutionary} {Computation} {Conference}},
    publisher = {ACM},
    author = {Mertan, Alican and Cheney, Nick},
    month = jul,
    year = {2023},
    pages = {174--183},
}

@inproceedings{cheney_difficulty_2016,
    address = {Cancun, Mexico},
    title = {On the {Difficulty} of {Co}-{Optimizing} {Morphology} and {Control} in {Evolved} {Virtual} {Creatures}},
    isbn = {978-0-262-33936-0},
    url = {https://www.mitpressjournals.org/doi/abs/10.1162/978-0-262-33936-0-ch042},
    doi = {10.7551/978-0-262-33936-0-ch042},
    language = {en},
    urldate = {2021-09-17},
    booktitle = {Proceedings of the {Artificial} {Life} {Conference} 2016},
    publisher = {MIT Press},
    author = {Cheney, Nicholas and Bongard, Josh and Sunspiral, Vytas and Lipson, Hod},
    year = {2016},
    pages = {226--233},
}

@article{cheney_scalable_2018,
    title = {Scalable co-optimization of morphology and control in embodied machines},
    volume = {15},
    issn = {1742-5689, 1742-5662},
    url = {https://royalsocietypublishing.org/doi/10.1098/rsif.2017.0937},
    doi = {10.1098/rsif.2017.0937},
    language = {en},
    number = {143},
    urldate = {2021-08-24},
    journal = {Journal of The Royal Society Interface},
    author = {Cheney, Nick and Bongard, Josh and SunSpiral, Vytas and Lipson, Hod},
    month = jun,
    year = {2018},
    pages = {20170937},
}

@misc{mertan_controller_2025,
    title = {Controller {Distillation} {Reduces} {Fragile} {Brain}-{Body} {Co}-{Adaptation} and {Enables} {Migrations} in {MAP}-{Elites}},
    url = {http://arxiv.org/abs/2504.06523},
    doi = {10.1145/3712256.3726368},
    language = {en},
    urldate = {2025-04-25},
    author = {Mertan, Alican and Cheney, Nick},
    month = apr,
    year = {2025},
    note = {arXiv:2504.06523 [cs]},
}

@article{bhatia_evolution_2021,
    title = {Evolution {Gym}: {A} {Large}-{Scale} {Benchmark} for {Evolving} {Soft} {Robots}},
    language = {en},
    author = {Bhatia, Jagdeep Singh and Jackson, Holly and Tian, Yunsheng and Xu, Jie and Matusik, Wojciech},
    year = {2021},
    pages = {14},
}

@article{schmidt_age-fitness_2010,
    title = {Age-fitness pareto optimization},
    language = {en},
    author = {Schmidt, Michael D and Lipson, Hod},
    year = {2010},
    pages = {2},
}

@article{mouret_illuminating_2015,
    title = {Illuminating search spaces by mapping elites},
    url = {http://arxiv.org/abs/1504.04909},
    language = {en},
    urldate = {2021-09-10},
    journal = {arXiv:1504.04909 [cs, q-bio]},
    author = {Mouret, Jean-Baptiste and Clune, Jeff},
    month = apr,
    year = {2015},
    note = {arXiv: 1504.04909},
}

@article{gupta_embodied_2021,
    title = {Embodied intelligence via learning and evolution},
    volume = {12},
    copyright = {2021 The Author(s)},
    issn = {2041-1723},
    url = {https://www.nature.com/articles/s41467-021-25874-z},
    doi = {10.1038/s41467-021-25874-z},
    language = {en},
    number = {1},
    urldate = {2024-02-24},
    journal = {Nature Communications},
    author = {Gupta, Agrim and Savarese, Silvio and Ganguli, Surya and Fei-Fei, Li},
    month = oct,
    year = {2021},
    note = {Number: 1
Publisher: Nature Publishing Group},
    pages = {5721},
}

@inproceedings{eiben_if_2020,
    address = {Cancún Mexico},
    title = {If it evolves it needs to learn},
    isbn = {978-1-4503-7127-8},
    url = {https://dl.acm.org/doi/10.1145/3377929.3398151},
    doi = {10.1145/3377929.3398151},
    language = {en},
    urldate = {2022-04-22},
    booktitle = {Proceedings of the 2020 {Genetic} and {Evolutionary} {Computation} {Conference} {Companion}},
    publisher = {ACM},
    author = {Eiben, A. E. and Hart, Emma},
    month = jul,
    year = {2020},
    pages = {1383--1384},
}

@inproceedings{le_goff_improving_2024,
    address = {Melbourne VIC Australia},
    title = {Improving {Efficiency} of {Evolving} {Robot} {Designs} via {Self}-{Adaptive} {Learning} {Cycles} and an {Asynchronous} {Architecture}},
    isbn = {9798400704956},
    url = {https://dl.acm.org/doi/10.1145/3638530.3664116},
    doi = {10.1145/3638530.3664116},
    language = {en},
    urldate = {2025-04-27},
    booktitle = {Proceedings of the {Genetic} and {Evolutionary} {Computation} {Conference} {Companion}},
    publisher = {ACM},
    author = {Le Goff, Leni and Hart, Emma},
    month = jul,
    year = {2024},
    pages = {1607--1615},
}

@misc{strgar_evolution_2024,
    title = {Evolution and learning in differentiable robots},
    url = {http://arxiv.org/abs/2405.14712},
    doi = {10.48550/arXiv.2405.14712},
    language = {en},
    urldate = {2025-04-27},
    publisher = {arXiv},
    author = {Strgar, Luke and Matthews, David and Hummer, Tyler and Kriegman, Sam},
    month = may,
    year = {2024},
    note = {arXiv:2405.14712 [cs]},
}

@article{pigozzi_evolving_2022,
    title = {Evolving {Modular} {Soft} {Robots} without {Explicit} {Inter}-{Module} {Communication} using {Local} {Self}-{Attention}},
    url = {http://arxiv.org/abs/2204.06481},
    doi = {10.1145/3512290.3528762},
    language = {en},
    urldate = {2022-04-22},
    journal = {arXiv:2204.06481 [cs]},
    author = {Pigozzi, Federico and Tang, Yujin and Medvet, Eric and Ha, David},
    month = apr,
    year = {2022},
    note = {arXiv: 2204.06481},
}

@article{lieberman_first-mover_1988,
    title = {First-mover advantages},
    volume = {9},
    copyright = {Copyright © 1988 John Wiley \& Sons, Ltd.},
    issn = {1097-0266},
    url = {https://onlinelibrary.wiley.com/doi/abs/10.1002/smj.4250090706},
    doi = {10.1002/smj.4250090706},
    language = {en},
    number = {S1},
    urldate = {2025-04-27},
    journal = {Strategic Management Journal},
    author = {Lieberman, Marvin B. and Montgomery, David B.},
    year = {1988},
    note = {\_eprint: https://onlinelibrary.wiley.com/doi/pdf/10.1002/smj.4250090706},
    pages = {41--58},
}

@article{liu_embodied_2025,
    title = {Embodied {Intelligence}: {A} {Synergy} of {Morphology}, {Action}, {Perception} and {Learning}},
    volume = {57},
    issn = {0360-0300, 1557-7341},
    shorttitle = {Embodied {Intelligence}},
    url = {https://dl.acm.org/doi/10.1145/3717059},
    doi = {10.1145/3717059},
    language = {en},
    number = {7},
    urldate = {2025-05-08},
    journal = {ACM Computing Surveys},
    author = {Liu, Huaping and Guo, Di and Cangelosi, Angelo},
    month = jul,
    year = {2025},
    pages = {1--36},
}

@misc{wang2025brainbody, 
    author = {Yuxing Wang}, 
    title = {Brain-Body Co-Design for Embodied Agents: A Survey of Neural Approaches}, 
    year = {2025}, 
    publisher = {GitHub}, 
    journal = {GitHub repository}, 
    howpublished = {\url{https://github.com/Yuxing-Wang-THU/SurveyBrainBody}}, 
}

@article{hiller_automatic_2012,
    title = {Automatic {Design} and {Manufacture} of {Soft} {Robots}},
    volume = {28},
    issn = {1552-3098, 1941-0468},
    url = {http://ieeexplore.ieee.org/document/6096440/},
    doi = {10.1109/TRO.2011.2172702},
    language = {en},
    number = {2},
    urldate = {2021-08-29},
    journal = {IEEE Transactions on Robotics},
    author = {Hiller, Jonathan and Lipson, Hod},
    month = apr,
    year = {2012},
    pages = {457--466},
}

@book{pfeifer_how_2007,
    address = {Cambridge, Massachusetts},
    title = {How the body shapes the way we think: a new view of intelligence},
    isbn = {978-0-262-16239-5},
    shorttitle = {How the body shapes the way we think},
    language = {en},
    publisher = {MIT Press},
    author = {Pfeifer, Rolf and Bongard, Josh},
    year = {2007},
}

@inproceedings{ying2019bench,
  title={Nas-bench-101: Towards reproducible neural architecture search},
  author={Ying, Chris and Klein, Aaron and Christiansen, Eric and Real, Esteban and Murphy, Kevin and Hutter, Frank},
  booktitle={International conference on machine learning},
  pages={7105--7114},
  year={2019},
  organization={PMLR}
}

@misc{dong_nas-bench-201_2020,
    title = {{NAS}-{Bench}-201: {Extending} the {Scope} of {Reproducible} {Neural} {Architecture} {Search}},
    shorttitle = {{NAS}-{Bench}-201},
    url = {http://arxiv.org/abs/2001.00326},
    doi = {10.48550/arXiv.2001.00326},
    language = {en},
    urldate = {2025-05-09},
    publisher = {arXiv},
    author = {Dong, Xuanyi and Yang, Yi},
    month = jan,
    year = {2020},
    note = {arXiv:2001.00326 [cs]},
}

@article{joachimczak_artificial_2016,
    title = {Artificial {Metamorphosis}: {Evolutionary} {Design} of {Transforming}, {Soft}-{Bodied} {Robots}},
    volume = {22},
    issn = {1064-5462, 1530-9185},
    shorttitle = {Artificial {Metamorphosis}},
    url = {https://direct.mit.edu/artl/article/22/3/271-298/2848},
    doi = {10.1162/ARTL_a_00207},
    language = {en},
    number = {3},
    urldate = {2021-10-22},
    journal = {Artificial Life},
    author = {Joachimczak, Michał and Suzuki, Reiji and Arita, Takaya},
    month = aug,
    year = {2016},
    pages = {271--298},
}

@article{mann1947test,
  title={On a test of whether one of two random variables is stochastically larger than the other},
  author={Mann, Henry B and Whitney, Donald R},
  journal={The annals of mathematical statistics},
  pages={50--60},
  year={1947},
  publisher={JSTOR}
}

@inproceedings{wang2023preco,
  title={Preco: Enhancing generalization in co-design of modular soft robots via brain-body pre-training},
  author={Wang, Yuxing and Wu, Shuang and Zhang, Tiantian and Chang, Yongzhe and Fu, Haobo and Fu, Qiang and Wang, Xueqian},
  booktitle={Conference on Robot Learning},
  pages={478--498},
  year={2023},
  organization={PMLR}
}

@inproceedings{song2024morphvae,
  title={MorphVAE: advancing morphological design of voxel-based soft robots with variational autoencoders},
  author={Song, Junru and Yang, Yang and Peng, Wei and Zhou, Weien and Wang, Feifei and Yao, Wen},
  booktitle={Proceedings of the AAAI Conference on Artificial Intelligence},
  volume={38},
  number={9},
  pages={10368--10376},
  year={2024}
}

@article{matthews2023efficient,
  title={Efficient automatic design of robots},
  author={Matthews, David and Spielberg, Andrew and Rus, Daniela and Kriegman, Sam and Bongard, Josh},
  journal={Proceedings of the National Academy of Sciences},
  volume={120},
  number={41},
  pages={e2305180120},
  year={2023},
  publisher={National Academy of Sciences}
}

@inproceedings{mersmann2011exploratory,
  title={Exploratory landscape analysis},
  author={Mersmann, Olaf and Bischl, Bernd and Trautmann, Heike and Preuss, Mike and Weihs, Claus and Rudolph, G{\"u}nter},
  booktitle={Proceedings of the 13th annual conference on Genetic and evolutionary computation},
  pages={829--836},
  year={2011}
}

@inproceedings{mersmann2010benchmarking,
  title={Benchmarking evolutionary algorithms: Towards exploratory landscape analysis},
  author={Mersmann, Olaf and Preuss, Mike and Trautmann, Heike},
  booktitle={International Conference on Parallel Problem Solving from Nature},
  pages={73--82},
  year={2010},
  organization={Springer}
}

@article{jeon2025convergent,
  title={Convergent functions, divergent forms},
  author={Jeon, Hyeonseong and Eftekhar, Ainaz and Walsman, Aaron and Zeng, Kuo-Hao and Farhadi, Ali and Krishna, Ranjay},
  journal={arXiv preprint arXiv:2505.21665},
  year={2025}
}

@article{nordmoen_map-elites_2021,
    title = {{MAP}-{Elites} {Enables} {Powerful} {Stepping} {Stones} and {Diversity} for {Modular} {Robotics}},
    volume = {8},
    issn = {2296-9144},
    url = {https://www.frontiersin.org/articles/10.3389/frobt.2021.639173},
    doi = {10.3389/frobt.2021.639173},
    language = {English},
    urldate = {2024-03-26},
    journal = {Frontiers in Robotics and AI},
    publisher = {Frontiers},
    author = {Nordmoen, Jørgen and Veenstra, Frank and Ellefsen, Kai Olav and Glette, Kyrre},
    month = apr,
    year = {2021},
}

@incollection{lehman2011novelty,
  title={Novelty search and the problem with objectives},
  author={Lehman, Joel and Stanley, Kenneth O},
  booktitle={Genetic programming theory and practice IX},
  pages={37--56},
  year={2011},
  publisher={Springer}
}

@article{lehman_abandoning_2011,
    title = {Abandoning {Objectives}: {Evolution} {Through} the {Search} for {Novelty} {Alone}},
    volume = {19},
    issn = {1063-6560, 1530-9304},
    shorttitle = {Abandoning {Objectives}},
    url = {https://direct.mit.edu/evco/article/19/2/189-223/1365},
    doi = {10.1162/EVCO_a_00025},
    language = {en},
    number = {2},
    urldate = {2021-09-10},
    journal = {Evolutionary Computation},
    author = {Lehman, Joel and Stanley, Kenneth O.},
    month = jun,
    year = {2011},
    pages = {189--223},
}

@inproceedings{nguyen2015innovation,
  title={Innovation engines: Automated creativity and improved stochastic optimization via deep learning},
  author={Nguyen, Anh Mai and Yosinski, Jason and Clune, Jeff},
  booktitle={Proceedings of the 2015 annual conference on genetic and evolutionary computation},
  pages={959--966},
  year={2015}
}

@inproceedings{templier2025time,
  title={Time to Play: Simulating Early-Life Animal Dynamics Enhances Robotics Locomotion Discovery},
  author={Templier, Paul and Janmohamed, Hannah and Labonte, David and Cully, Antoine},
  booktitle={Artificial Life Conference Proceedings 37},
  volume={2025},
  number={1},
  pages={71},
  year={2025},
  organization={MIT Press One Rogers Street, Cambridge, MA 02142-1209, USA journals-info~…}
}

@article{ziemke2002neuromodulation,
  title={Neuromodulation of reactive sensorimotor mappings as a short-term memory mechanism in delayed response tasks},
  author={Ziemke, Tom and Thieme, Mikael},
  journal={Adaptive Behavior},
  volume={10},
  number={3-4},
  pages={185--199},
  year={2002},
  publisher={Sage Publications}
}

@inproceedings{maris1996exploiting,
  title={Exploiting physical constraints: heap formation through behavioral error in a group of robots},
  author={Maris, Marinus and Boeckhorst, Ren{\'e}},
  booktitle={Proceedings of IEEE/RSJ international conference on intelligent robots and systems. IROS'96},
  volume={3},
  pages={1655--1660},
  year={1996},
  organization={IEEE}
}

@book{braitenberg1986vehicles,
  title={Vehicles: Experiments in synthetic psychology},
  author={Braitenberg, Valentino},
  year={1986},
  publisher={MIT press}
}

@article{mcgeer1990passive,
  title={Passive dynamic walking},
  author={McGeer, Tad},
  journal={The international journal of robotics research},
  volume={9},
  number={2},
  pages={62--82},
  year={1990},
  publisher={Sage Publications Sage CA: Thousand Oaks, CA}
}

@article{beal2006passive,
  title={Passive propulsion in vortex wakes},
  author={Beal, David N and Hover, Franz S and Triantafyllou, Michael S and Liao, James C and Lauder, George V},
  journal={Journal of fluid mechanics},
  volume={549},
  pages={385--402},
  year={2006},
  publisher={Cambridge University Press}
}

@article{hauser2011towards,
  title={Towards a theoretical foundation for morphological computation with compliant bodies},
  author={Hauser, Helmut and Ijspeert, Auke J and F{\"u}chslin, Rudolf M and Pfeifer, Rolf and Maass, Wolfgang},
  journal={Biological cybernetics},
  volume={105},
  pages={355--370},
  year={2011},
  publisher={Springer}
}

@book{lakoff2008metaphors,
  title={Metaphors we live by},
  author={Lakoff, George and Johnson, Mark},
  year={2008},
  publisher={University of Chicago press}
}

@article{ferigo2025totipotent,
  title={Totipotent neural controllers for modular soft robots: Achieving specialization in body--brain co-evolution through Hebbian learning},
  author={Ferigo, Andrea and Iacca, Giovanni and Medvet, Eric and Nadizar, Giorgia},
  journal={Neurocomputing},
  volume={614},
  pages={128811},
  year={2025},
  publisher={Elsevier}
}

@inproceedings{yu2018towards,
      title={Towards Sample Efficient Reinforcement Learning.},
      author={Yu, Yang},
      booktitle={IJCAI},
      pages={5739--5743},
      year={2018}
    }

@article{xu2022accelerated,
      title={Accelerated policy learning with parallel differentiable simulation},
      author={Xu, Jie and Makoviychuk, Viktor and Narang, Yashraj and Ramos, Fabio and Matusik, Wojciech and Garg, Animesh and Macklin, Miles},
      journal={arXiv preprint arXiv:2204.07137},
      year={2022}
    }

@misc{gupta_metamorph_2022,
    title = {{MetaMorph}: {Learning} {Universal} {Controllers} with {Transformers}},
    shorttitle = {{MetaMorph}},
    url = {http://arxiv.org/abs/2203.11931},
    language = {en},
    urldate = {2024-01-09},
    publisher = {arXiv},
    author = {Gupta, Agrim and Fan, Linxi and Ganguli, Surya and Fei-Fei, Li},
    month = mar,
    year = {2022},
    note = {arXiv:2203.11931 [cs]},
}

@article{wang_nervenet_2018,
    title = {{NERVENET}: {LEARNING} {STRUCTURED} {POLICY} {WITH} {GRAPH} {NEURAL} {NETWORKS}},
    language = {en},
    author = {Wang, Tingwu and Liao, Renjie and Fidler, Sanja},
    year = {2018},
}

@article{zhao_robogrammar_2020,
    title = {{RoboGrammar}: graph grammar for terrain-optimized robot design},
    volume = {39},
    issn = {0730-0301, 1557-7368},
    shorttitle = {{RoboGrammar}},
    url = {https://dl.acm.org/doi/10.1145/3414685.3417831},
    doi = {10.1145/3414685.3417831},
    language = {en},
    number = {6},
    urldate = {2022-03-15},
    journal = {ACM Transactions on Graphics},
    author = {Zhao, Allan and Xu, Jie and Konaković-Luković, Mina and Hughes, Josephine and Spielberg, Andrew and Rus, Daniela and Matusik, Wojciech},
    month = dec,
    year = {2020},
    pages = {1--16},
}

@inproceedings{medvet2022impact,
  title={Impact of morphology variations on evolved neural controllers for modular robots},
  author={Medvet, Eric and Rusin, Francesco},
  booktitle={Italian Workshop on Artificial Life and Evolutionary Computation},
  pages={266--277},
  year={2022},
  organization={Springer}
}

@article{luo2023lamarck,
  title={Lamarck's Revenge: Inheritance of Learned Traits Can Make Robot Evolution Better},
  author={Luo, Jie and Miras, Karine and Tomczak, Jakub and Eiben, Agoston E},
  journal={arXiv preprint arXiv:2309.13099},
  year={2023}
}

@inproceedings{mertan2025evolutionary,
  title={Evolutionary Brain-Body Co-Optimization Consistently Fails to Select for Morphological Potential},
  author={Mertan, Alican and Cheney, Nick},
  booktitle={Artificial Life Conference Proceedings 37},
  volume={2025},
  number={1},
  pages={63},
  year={2025},
  organization={MIT Press One Rogers Street, Cambridge, MA 02142-1209, USA journals-info~…}
}

@inproceedings{ferigo2022entanglement,
  title={On the entanglement between evolvability and fitness: An experimental study on voxel-based soft robots},
  author={Ferigo, Andrea and Soros, LB and Medvet, Eric and Iacca, Giovanni},
  booktitle={Artificial Life Conference Proceedings 34},
  volume={2022},
  number={1},
  pages={15},
  year={2022},
  organization={MIT press One Rogers Street, Cambridge, MA 02142-1209, USA journals-info~…}
}

@inproceedings{pigozzi2023morphology,
  title={How the morphology encoding influences the learning ability in body-brain co-optimization},
  author={Pigozzi, Federico and Camerota Verd{\`u}, Federico Julian and Medvet, Eric},
  booktitle={Proceedings of the Genetic and Evolutionary Computation Conference},
  pages={1045--1054},
  year={2023}
}

@inproceedings{mertan2025morphological,
  title={Morphological Cognition: Classifying MNIST Digits Through Morphological Computation Alone},
  author={Mertan, Alican and Cheney, Nick},
  booktitle={Artificial Life Conference Proceedings 37},
  volume={2025},
  number={1},
  pages={64},
  year={2025},
  organization={MIT Press One Rogers Street, Cambridge, MA 02142-1209, USA journals-info~…}
}

@inproceedings{mertan2024no,
  title={No-brainer: Morphological computation driven adaptive behavior in soft robots},
  author={Mertan, Alican and Cheney, Nick},
  booktitle={International Conference on Simulation of Adaptive Behavior},
  pages={81--92},
  year={2024},
  organization={Springer}
}

@inproceedings{mertan2024towards,
  title={Towards multi-morphology controllers with diversity and knowledge distillation},
  author={Mertan, Alican and Cheney, Nick},
  booktitle={Proceedings of the Genetic and Evolutionary Computation Conference},
  pages={367--376},
  year={2024}
}

\end{document}